\documentclass[runningheads]{llncs}
\usepackage{graphicx}

\usepackage{tikz}
\usepackage{comment}
\usepackage{amsmath,amssymb} 
\usepackage{color,xcolor,colortbl}
\usepackage{subcaption}
\usepackage{hyperref}
\usepackage{orcidlink}
\usepackage{pifont}

\usepackage[accsupp]{axessibility}  

\newcommand{\cmark}{\text{\ding{51}}}%
\newcommand{\xmark}{\text{\ding{55}}}%

\begin{document}
\pagestyle{headings}
\mainmatter
\def\ECCVSubNumber{360}  

\title{Reversing Image Signal Processors by Reverse Style Transferring} 

\titlerunning{Reversing ISP by Reverse Style Transferring}
%
\author{Furkan K{\i}nl{\i}\textsuperscript{1}*
\orcidlink{0000-0002-9192-6583} 
\and
Bar{\i}\c{s} \"{O}zcan\textsuperscript{2}
\orcidlink{0000-0001-8598-1239} 
\and
Furkan K{\i}ra\c{c}\textsuperscript{3}
\orcidlink{0000-0001-9177-0489}
}
\authorrunning{F. K{\i}nl{\i} et al.}
%
\institute{\"{O}zye\u{g}in University, \.{I}stanbul, Turkey \\
\email{\{\textsuperscript{1,*}furkan.kinli, \textsuperscript{3}furkan.kirac\}@ozyegin.edu.tr}
\email{\textsuperscript{2}baris.ozcan.10097@ozu.edu.tr}}
\maketitle

\begin{abstract}
RAW image datasets are more suitable than the standard RGB image datasets for the ill-posed inverse problems in low-level vision, but not common in the literature. There are also a few studies to focus on mapping sRGB images to RAW format. Mapping from sRGB to RAW format could be a relevant domain for reverse style transferring since the task is an ill-posed reversing problem. In this study, we seek an answer to the question: Can the ISP operations be modeled as the style factor in an end-to-end learning pipeline? To investigate this idea, we propose a novel architecture, namely \textit{RST-ISP-Net}, for learning to reverse the ISP operations with the help of adaptive feature normalization. We formulate this problem as a reverse style transferring and mostly follow the practice used in the prior work. We have participated in the AIM Reversed ISP challenge with our proposed architecture. Results indicate that the idea of modeling disruptive or modifying factors as style is still valid, but further improvements are required to be competitive in such a challenge.

\keywords{Image Signal Processors, Reverse Style Transfer, sRGB-to-RAW Reconstruction}
\end{abstract}

\section{Introduction}
\footnote{*: Corresponding author}
Data-driven learning methods such as Convolutional Neural Networks (CNNs) achieved outstanding results in vision tasks such as object detection, image segmentation, or image classification. These methods require a huge amount of annotated images, which is increasingly available in recent years \cite{deng2009imagenet,lin2014microsoft}. Most of these datasets consist of standard RGB (sRGB) images, which are in-camera Image Signal Processor (ISP) dependent. An sRGB image can be obtained by feeding the RAW data acquired by the camera sensors to the ISP pipeline that outputs an image tailored for human perception. However, RAW images are found to be better suited for ill-posed low-level vision tasks such as denoising, HDR, or super-resolution \cite{conde2022model}. Unfortunately, there are very few RAW image datasets that are available, therefore deep learning-based methods were not fully utilized. The researchers often rely on synthetically-generated data for low-level vision tasks. However, deep learning-based denoising methods trained on such data outperform the hand-engineered methods in the recent datasets when tested on real raw images \cite{brooks2019unprocessing}. Unrealistic synthetic data  used for training the deep learning models are argued to be the limitation behind better results.

To improve the synthetic RAW image quality, recent studies aim to estimate the RAW image from the sRGB data \cite{brooks2019unprocessing,punnappurath2019learning,zamir2020cycleisp} when attacking low-level vision tasks. Reversing the ISP operations is an ill-posed inverse problem. In this context, we have approached this task as a reverse style transferring problem and focused on observing the limitations of our approach as an emerging idea. Therefore, we have participated in the AIM Reversed ISP challenge \cite{conde2022aim} to be able to compare with the other methods on a fair basis. This challenge aims to obtain a network design or a solution, which is capable of producing high-quality results with high fidelity with respect to the reference ground truth. Our method is one of the top novel solutions for the proposed problem of RAW image reconstruction, which is a novel inverse problem in low-level computer vision.

The following sections of this paper can be summarized as follows. Section \ref{sec:rworks} introduces the previous studies on both sRGB-to-RAW reconstruction and reverse style transferring. Section \ref{sec:proposed} presents our proposed methodology and the idea of modeling disruptive factors as style. Section \ref{sec:exps} explains the datasets used in the scope of the challenge, the experimental details, and the results obtained by our solution and the other studies. Section \ref{sec:concl} concludes the paper.

\section{Related Works}
\label{sec:rworks}
\subsection{Mapping from sRGB to RAW}

Despite the fact that the RAW image datasets are more suitable than the standard RGB image datasets for the ill-posed inverse problems in low-level vision such as denoising, demosaicking, and super-resolution and the lack of these kinds of datasets, there are a few studies to focus on mapping sRGB images to RAW format in the literature. \cite{brooks2019unprocessing} proposes a generic camera ISP model, which replaces five fundamental ISP operations by approximating each of them by an invertible and differentiable function. Recent studies \cite{punnappurath2019learning,zamir2020cycleisp,xing2021invertible} are more focused on learning-based approaches where the ISP operations are learned by neural networks in an end-to-end manner. \cite{zamir2020cycleisp} employs the cycle consistency idea to learn two-way translations (\textit{i.e.}, RAW-to-RGB and RGB-to-RAW) for the input images. \cite{xing2021invertible} proposes theISP architecture that mimics the camera ISP model by learning a single invertible neural network for two-way translations. Recently, \cite{conde2022model} presents a hybrid approach, which addresses the issues that the previous approaches face. It is based on \cite{brooks2019unprocessing}, yet builds a more flexible and interpretable ISP pipeline by combining model-based and learning-based approaches. In this study, rather than seeking the best approach for the ISP in the AIM Reversed ISP challenge, we mostly seek an answer to the question of whether the most simple version of modeling the style can model the ISP operations as the style factor in an end-to-end learning pipeline is possible or not. Moreover, there is another AIM challenge that focuses on learned smartphone ISP on mobile GPUs by using deep learning strategies, and the solutions for this topic can be found in the challenge report \cite{ignatov2022isp}.

\subsection{Reverse Style Transferring}

Style Transfer \cite{Gatys2015c,46163,huang2017adain} is a common term in deep learning literature, which is a many-to-many translation approach where a reference image can be translated into a target image without losing the main context, but by transferring its style information. From a different perspective, reverse style transfer is described in \cite{Kinli_2021_CVPR} as a many-to-one translation approach for eliminating undesired style information from the input. Any number of transformations applied to an image can be swept away by adaptively normalizing throughout the encoding part of the network, and the output with the pure style can be generated by the decoder in an end-to-end learning pipeline. Recently, \cite{Kinli_2022_CVPR} introduces the patch-wise contrastive learning-based approach for reverse style transferring in the filter removal task, and investigates the importance of distilling the semantic and style similarities among the signals (\textit{i.e.}, patches). The main motivation of this study is to apply this strategy in a more primitive form to another reverse problem (\textit{i.e.}, reversed ISP) in order to prove the idea of modeling disruptive or modifying factors as the style.

\section{Proposed Methodology}
\label{sec:proposed}

\begin{figure}[t]
   \begin{center}
     \includegraphics[width=\textwidth]{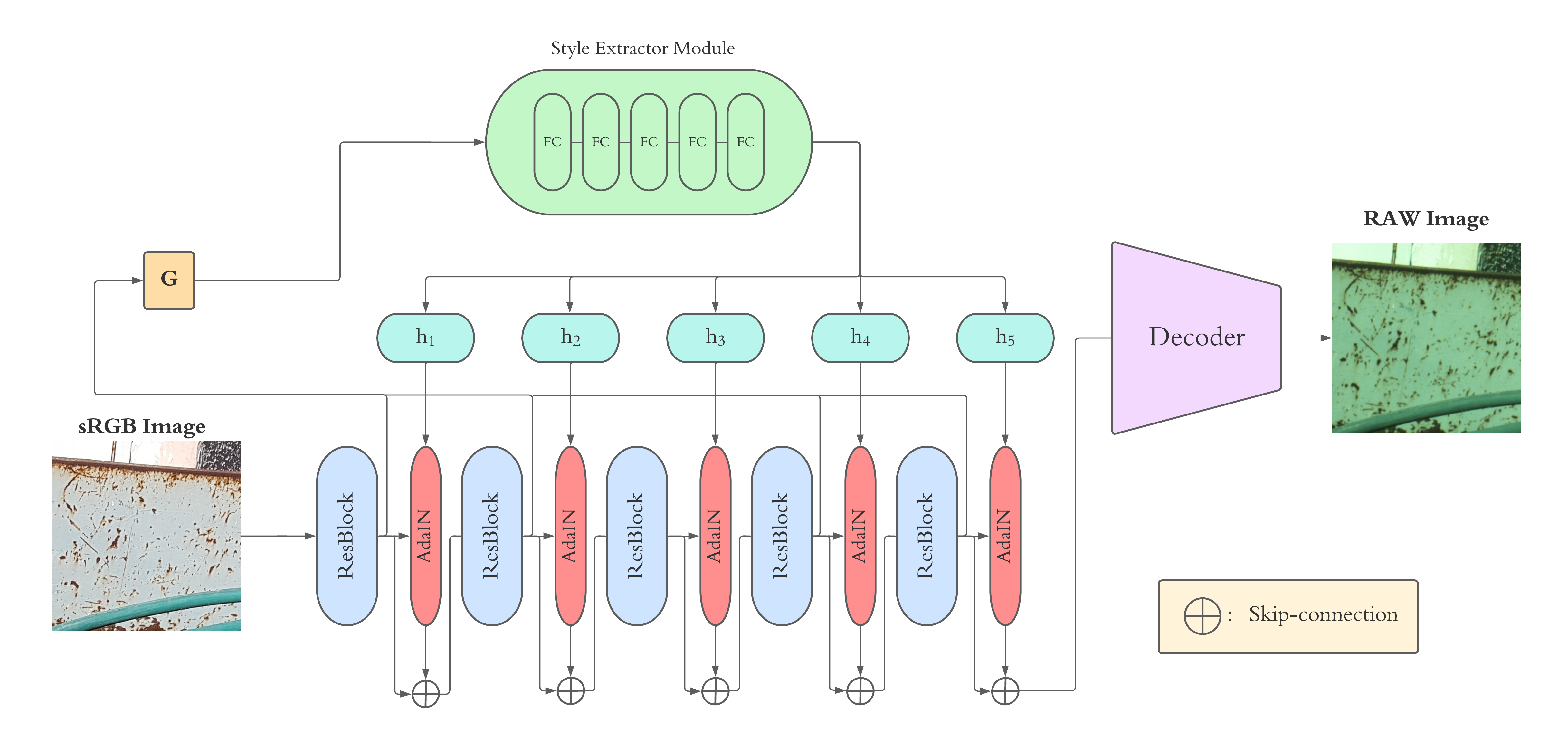}
      \caption{Overall architecture of our proposed strategy to sRGB-to-RAW reconstruction task.}  
      \label{fig:arch} 
      \end{center}
\end{figure}

We define the problem of reconstructing RAW images from sRGB input images as a reverse style transfer problem. Assuming the RAW format is the original version, any visual change related to the image signal processors (ISP) operations can be considered as an additional style factor injected into the original version. Following the idea in \cite{Kinli_2021_CVPR}, we can model the effects brought by the ISP's operations as the style factor, and remove these injected changes by directly reverting them back to their original style (\textit{i.e.}, RAW versions for this task). To achieve this, we propose a novel architecture, namely \textit{RST-ISP-Net}, for learning to reverse the ISP operations with the help of adaptive feature normalization for transferring the style information. Figure \ref{fig:arch} demonstrates the overall architecture of our proposed strategy for this task.

A given image $\mathbf{\Tilde{X}} \in \mathbb{R}^{H \times W \times 3}$ including a style information of the ISP operations applied by arbitrary transformation functions $\mathbf{T}(\cdot)$ is converted back to its original version $\mathbf{X} \in \mathbb{R}^{H \times W \times 3}$. This refers to $\mathbf{X}$ having the pure style and does not contain any additional style information injected. We formulate a style removal module $\mathbf{F}(\cdot)$, which is responsible for reverting the function of $\mathbf{T}(\cdot)$.

\begin{equation}
    \mathbf{X} = \mathbf{F}(\mathbf{\Tilde{X}})
\end{equation}
where $\mathbf{\Tilde{X}} = \mathbf{T}(\mathbf{X})$ and $\mathbf{T}(\cdot)$ is a general transformation function representing one or more transformations applied to $\mathbf{X}$ during the ISP. Note that finding $\mathbf{T}^{-1}(\cdot)$ is an ill-posed problem. 

Our proposed architecture has an encoder-decoder structure, and employs adaptive feature normalization between all layers. Style extractor module $f_{fc}$ is consist of five consecutive fully-connected layers to map the feature representations extracted by the Gram matrix to the latent space. We have $N$ fully-connected head layers attached to $f_{fc}$ where $N$ is the number of layer levels used in the encoder. These heads adapt the affine parameters so that the adaptive feature normalization modules in each level take them as the input and normalize the feature maps accordingly. This part of the architecture is responsible for reversing the style, in other words, removing the injected style factor from the input feature maps. This part can be formulated as follows

\begin{equation}
    y_{i} = h_{i}(f_{fc}(\mathbf{z}_{\mathbf{\Tilde{X}}})) 
\end{equation}
where $\mathbf{z}_{\mathbf{\Tilde{X}}}$ represents the feature representation of the input image $\mathbf{\Tilde{X}}$, $h_{i}(\cdot)$ is $i^{th}$ fully-connected head of style extractor module and $y_{i}$ represents the predicted mean and variance vectors for the corresponding normalization layer. In this study, we follow the literature in reverse style transferring and use Adaptive Instance Normalization (AdaIN) \cite{huang2017adain} as the normalization layer. We can formulate AdaIN as follows

\begin{equation}
    \text{AdaIN}(x, y) = \sigma(y) \left(\frac{x - \mu(x)}{\sigma(x)}\right) + \mu(y)
\end{equation}
where $\mu$ stands for the mean and $\sigma$ for the variance of the content image $x$ and the style input $y$. 

The encoder of RST-ISP-Net is composed of 5 residual blocks, and each of them has a specific AdaIN layer to normalize the feature maps in each level with its corresponding affine parameters extracted by the previous module. Following \cite{Kinli_2021_CVPR}, we include skip-connections between the residual blocks to be able to preserve the style information related to the pure style. Apart from the previous studies, we employ the Gram matrix for expressing the style information of the feature maps via feature correlations, instead of using a particular layer outputs of pre-trained neural networks (\textit{e.g.}, VGG16 \cite{Simonyan15}). The Gram matrix $\mathbf{G}^l \in \mathbb{R}^{K \times K}$ for each layer level $l$ represents the inner product between the features mapped by the corresponding encoder layer $\mathbf{E}^l$, and sends its output to the fully-connected head layer $\mathbf{H}^l$. We assume that the output of the last encoder layer should include no external style information and all additional injected style information is discarded. At this point, we can feed the feature representations without external style information into the decoder part to generate the RAW output image. The decoder also contains 4 common-typed residual blocks with PixelShuffle \cite{Shi_2016_CVPR} to interpolate the feature maps to the original resolution. Moreover, we apply discriminative regularization \cite{lamb2016discriminative} by Wavelet-based discriminators \cite{wang2020multi} to our network to avoid the blurring effect on the outputs and to preserve the high-frequency details for RAW format.

The objective function for our proposed strategy is composed of three main components, which are MS-SSIM loss \cite{1292216}, TV loss \cite{liu2019image}, and adversarial loss for providing the discriminative regularization. We did not include the auxiliary classification loss used in \cite{Kinli_2021_CVPR} since the improvement rate in the performance is not significant with respect to the computational burden increased. The general formula of our adversarial training can be seen in Equation \ref{eq:objfn}.

\begin{equation}
    \mathcal{L} = \lambda_{SSIM}\mathcal{L}_{SSIM} + \lambda_{TV}\mathcal{L}_{TV} + \lambda_{adv}\mathcal{L}_{adv} + \lambda_{gp}\mathcal{L}_{gp}
    \label{eq:objfn}
\end{equation}
where $\mathcal{L}_{gp}$ represents the gradient penalty applied on the discriminator whose weight $\lambda_{gp}$ is set to 10.

\section{Experiments and Results}
\label{sec:exps}

\begin{table}[t]
    \centering
    \resizebox{\textwidth}{!}{
        \begin{tabular}{c|c|c|c|c|c|c|c|c|c|c}
            Dataset & Input &  Opt. & LR & Epochs & Batch size & Ensemble & Framework & \# Params. (M) & Runtime (ms) & GPU  \\
            \hline
            S7 & (504,504,3) & Adam & 1e-4 & 101 & 8 & No & PyTorch & 86.3 Million & 5.45 on GPU & RTX 2080Ti \\
            P20 & (496,496,3) & Adam & 1e-4 & 101 & 8 & No & PyTorch & 86.3 Million & 5.54 on GPU & RTX 2080Ti
        \end{tabular}
    }
    \caption{Hyper-parameters used in our experiments.}
    \label{tab:params}
\end{table}

\subsection{Datasets}

In this study, we have used two datasets in our experiments, which are Samsung S7 DeepISP Dataset \cite{schwartz2018deepisp} and ETH Huawei P20 Dataset \cite{ignatov2020replacing}. The first is a dataset of real-world images that contains different scenes captured by a Samsung S7 rear camera. To ensure to avoid camera movement, a special Android application is used during the collection of images. Although the whole dataset contains a total of 110 scenes were captured and split into 90, 10
and 10 for the training, validation, and test sets, respectively, and the images are in 12Mpx resolution, we have only used the set of training images given by the AIM Reversed ISP challenge. Next, we have used the samples from the Zurich RAW-to-RGB dataset (\textit{i.e.}, called ETH Huawei P20 Dataset in the original challenge repository), which is a large-scale dataset consisting of 20K photos collected by Huawei P20 smartphone with their RAW versions. The data was collected over several weeks in a variety of places and in various illumination and weather conditions. Similar to the first one, we have only used the set of certain images given by the AIM Reversed ISP challenge track. 

\subsection{Experimental Details}

We have used the given input images without applying any pre-processing, thus the input size is $504 \times 504 \times 3$ for S7 dataset and $496 \times 496 \times 3$ for P20 dataset. We picked the Adam optimizer \cite{DBLP:journals/corr/KingmaB14} for our experiments with the learning rate of $1e-4$ for the generator and $4e-4$ for the discriminator. We have trained our model until proper convergence on the generator loss (\textit{i.e.}, 101 epochs for S7 dataset, 52 epochs for P20 dataset). We did not use any extra data in addition to the given training data, and also did not employ any ensembling strategy for our solution. The main reason is to be able to observe the baseline performance of the aforementioned approach, and the main aim is to seek some clues for improving this idea. The architecture used in our solution has 86.3M parameters, mostly due to the style projectors for each residual block. We have trained our method from scratch for all components. We did not use any additional data in addition to the provided training data. Table \ref{tab:params} summarizes the hyper-parameters used in our experiments. During testing, we did not apply any preprocessing to the given test images. Following the evaluation methods given in AIM Reversed ISP challenge \cite{conde2022aim}, we have measured the performance of our proposed architecture with two metrics (\textit{i.e.}, PSNR and SSIM). To visualize the RAW outputs, we have used the script\footnote{\url{}https://github.com/mv-lab/AISP} given by the AIM Reversed ISP challenge organizers.

\begin{table}[!t]
    \centering
    \caption{AIM Reversed ISP Challenge Benchmark \cite{conde2022aim}. Teams are ranked based on their performance on \underline{Test2}, an internal test set to evaluate the generalization capabilities and robustness of the proposed solutions. \underline{Test1} is a public test set provided to the participants as performance guidance. We report the standard metrics PSNR and SSIM. ED indicates the use of extra datasets besides the provided challenge datasets, ENS stands for if the solution is an ensemble of multiple models.}
    \label{tab:ablation}
    \resizebox{\linewidth}{!}{
        \begin{tabular}{l||c|c||c|c||c|c||c|c||c|c}
            \hline\noalign{\smallskip} & &
            & \multicolumn{4}{c ||}{\textbf{Track 1 (Samsung S6)}} & \multicolumn{4}{c}{\textbf{Track 2 (Huawei P20)}} \\
            Team & & & \multicolumn{2}{c||}{Test1} & \multicolumn{2}{c||}{Test2} & \multicolumn{2}{c||}{Test1} & \multicolumn{2}{c}{Test2} \\
             name & ED & ENS & PSNR$\uparrow$ & SSIM$\uparrow$ & PSNR$\uparrow$ & SSIM$\uparrow$ & PSNR$\uparrow$ & SSIM$\uparrow$ & PSNR$\uparrow$ & SSIM$\uparrow$ \\
            \hline
            NOAHTCV	 & \xmark & \xmark   & 31.86 & 0.83 & 32.69 & 0.88 & 38.38 & 0.93 & 35.77 & 0.92 \\
            MiAlgo	 & \xmark & \xmark   & 31.39 & 0.82 & 30.73 & 0.80 & 40.06 & 0.93 & 35.41 & 0.91 \\
            CASIALCVG	& \cmark & \cmark   & 30.19 & 0.81 & 31.47 & 0.86 & 37.58 & 0.93 & 33.99 & 0.92  \\
            HIT-IIL	 & \xmark & \xmark   & 29.12 & 0.80 & 29.98 & 0.87 & 36.53 & 0.91 & 34.07 & 0.90 \\
            CS2U   & \cmark & \cmark   & 29.13 & 0.79 & 29.95 & 0.84 & -     & - & -     & -   \\
            SenseBrains	& \xmark & \cmark   & 28.36 & 0.80 & 30.08 & 0.86 & 35.47 & 0.92 & 32.63 & 0.91 \\
            PixelJump   & \xmark & \cmark   & 28.15 & 0.80 & \textit{n/a} & \textit{n/a} & -     & - & -     & -   \\
            HiImage	   & \xmark & \xmark   & 27.96 & 0.79 & \textit{n/a} & \textit{n/a} & 34.40  & 0.94 & 32.13   & 0.90  \\
            0noise	    & \xmark & \xmark   & 27.67 & 0.79 & 29.81 & 0.87 & 33.68 & 0.90 & 31.83 & 0.89 \\
            OzU VGL	\textbf{(Ours)} & \xmark & \xmark  & 27.89 & 0.79 & 28.83 & 0.83 & 32.72 & 0.87 & 30.69 & 0.86 \\
            CVIP	  & \xmark & \xmark  & 27.85 & 0.80 & 29.50 & 0.86 & -     & - & -     & -   \\
            \hline
        \end{tabular}
    }
\end{table}

We have used Python language, and PyTorch DL framework \cite{NEURIPS2019_9015} for DL modeling and training/testing. Our experiments have been done on $2 \times$ NVIDIA RTX 2080Ti GPUs. The batch size is set to 8. Downloading and preparing the dataset for training and validation took approximately 1 day, and completing the implementation by using the code of the baseline study took approximately 2-3 days. A single experiment for training has been completed in approximately 2 days. The testing process has been completed in only 1 minute for all instances of the dataset. Run-time at test per image has been measured as 5 ms. for both datasets. The source code will be given for the camera-ready submission.

\captionsetup[subfigure]{font=small, labelformat=empty}
\begin{figure}[!t]
        \centering
        \begin{subfigure}[b]{0.11\textwidth}
                \includegraphics[width=\textwidth]{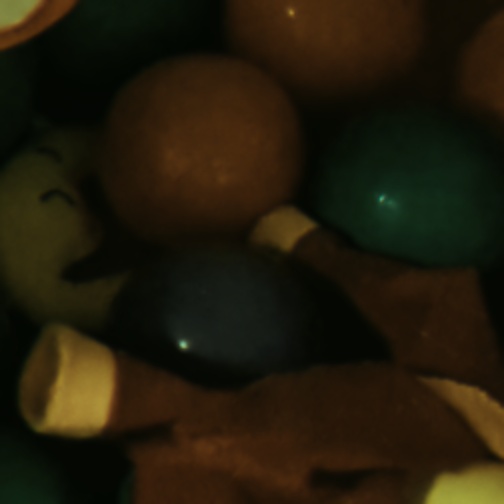}
                \includegraphics[width=\textwidth]{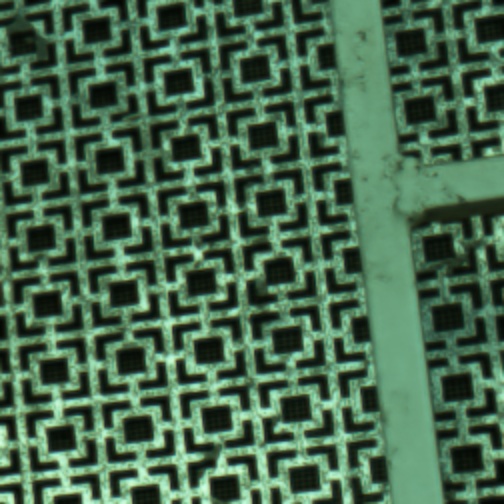}
                \includegraphics[width=\textwidth]{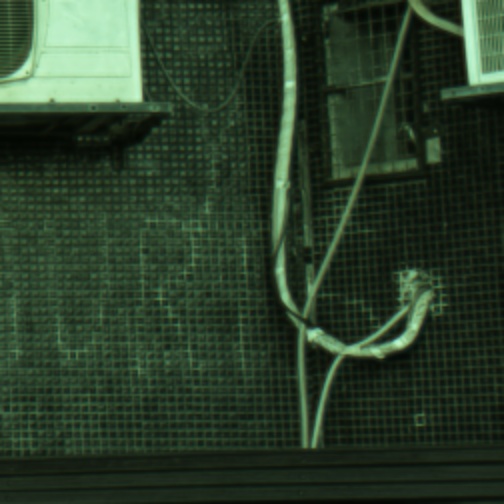}
                \includegraphics[width=\textwidth]{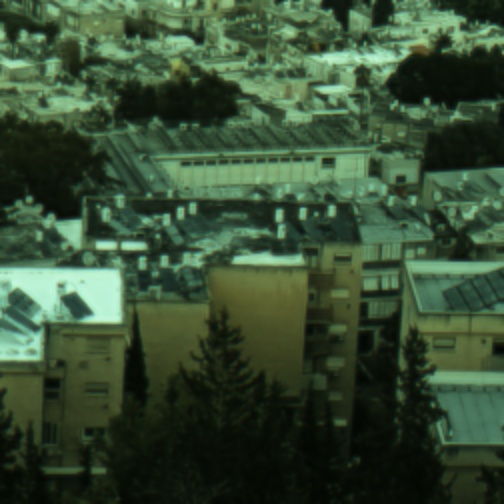}
                \includegraphics[width=\textwidth]{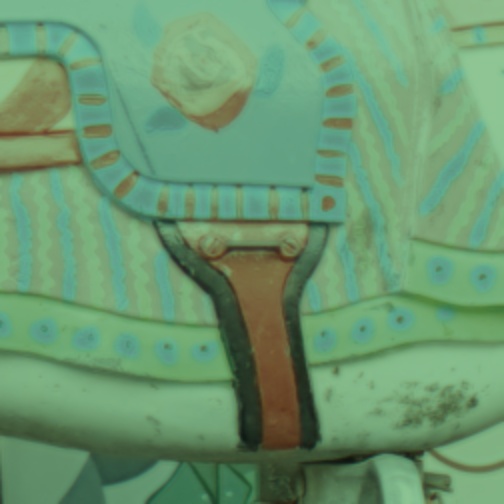}
                \includegraphics[width=\textwidth]{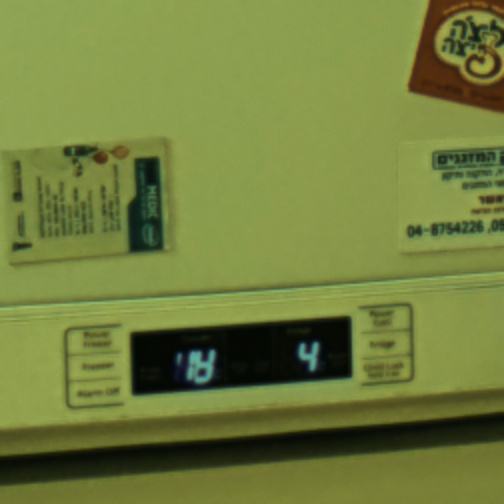}
                \includegraphics[width=\textwidth]{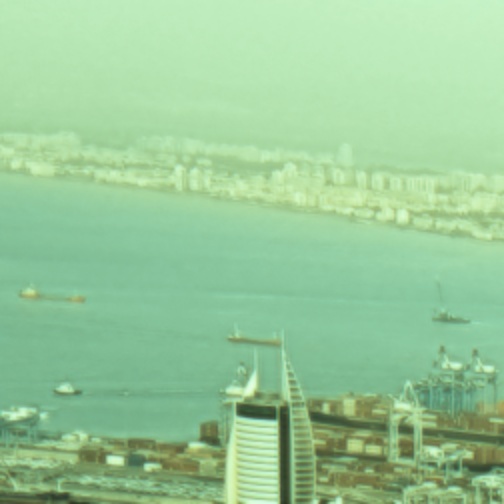}
                \label{fig:0noise}
                \caption{0noise}
        \end{subfigure}  
        \begin{subfigure}[b]{0.11\textwidth}
                \includegraphics[width=\textwidth]{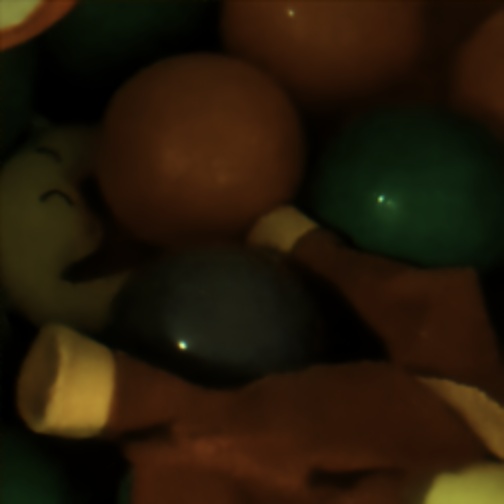}
                \includegraphics[width=\textwidth]{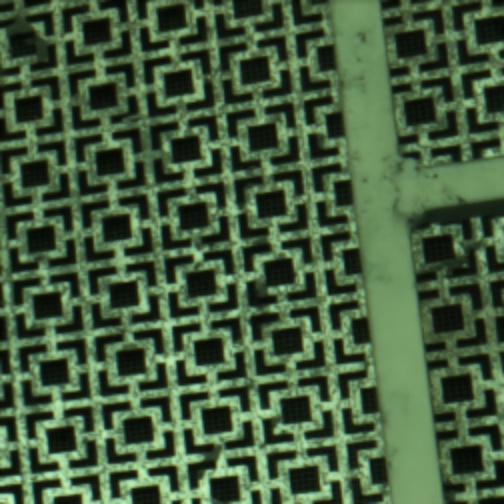}
                \includegraphics[width=\textwidth]{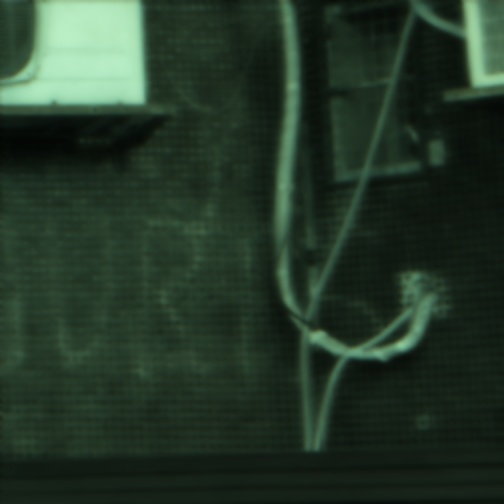}
                \includegraphics[width=\textwidth]{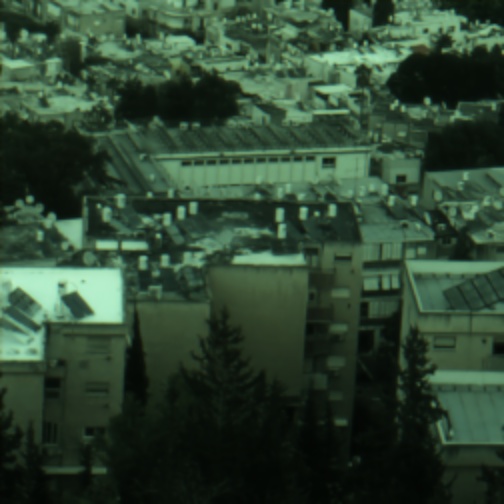}
                \includegraphics[width=\textwidth]{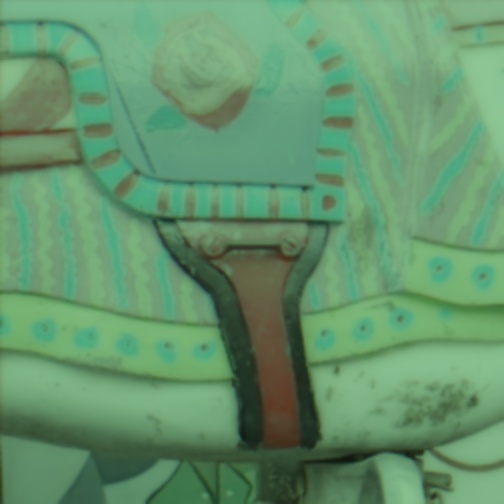}
                \includegraphics[width=\textwidth]{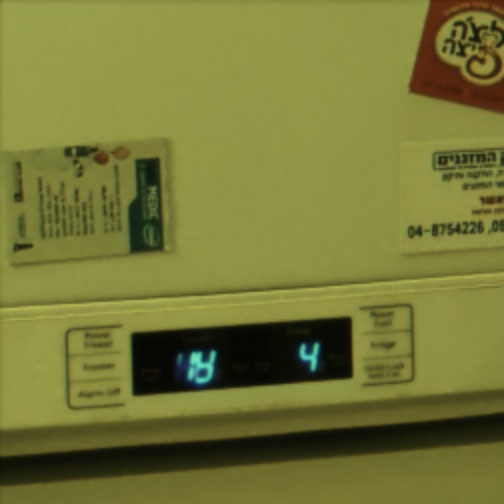}
                \includegraphics[width=\textwidth]{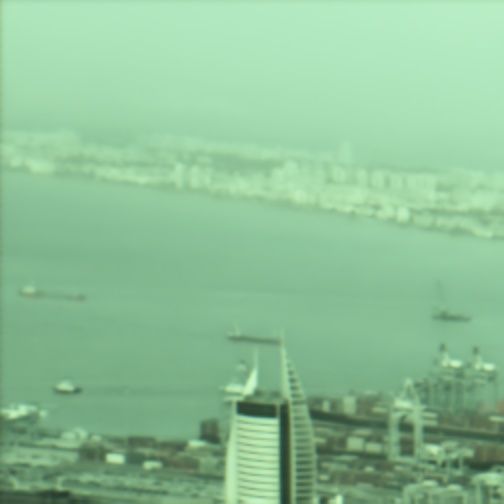}
                \label{fig:casia}
                \caption{CASIA}
        \end{subfigure}       
        \begin{subfigure}[b]{0.11\textwidth}
              \includegraphics[width=\textwidth]{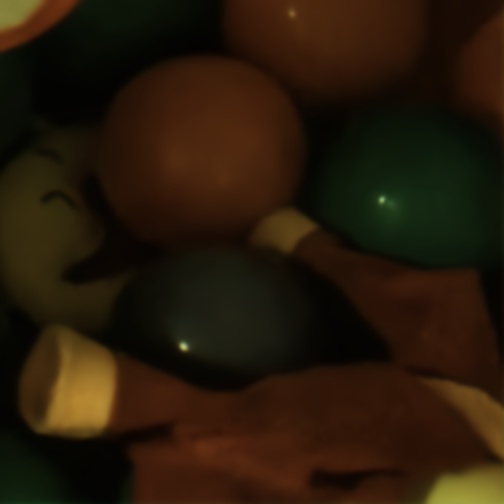}
                \includegraphics[width=\textwidth]{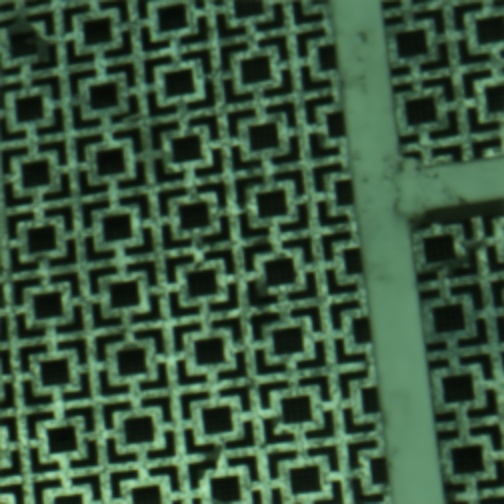}
                \includegraphics[width=\textwidth]{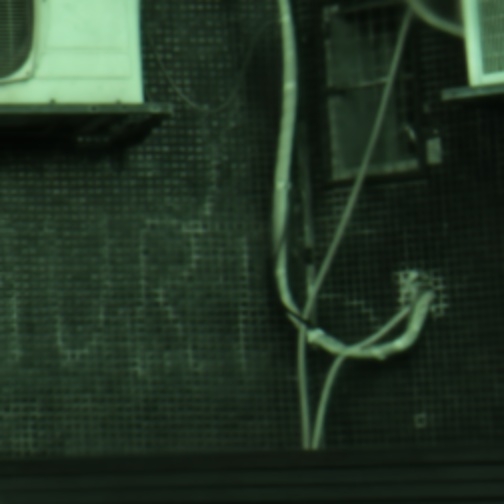}
                \includegraphics[width=\textwidth]{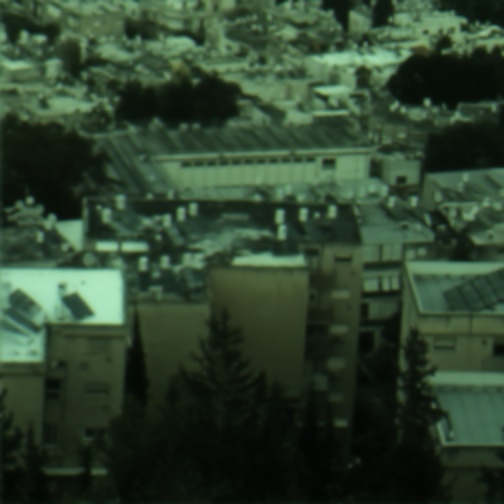}
                \includegraphics[width=\textwidth]{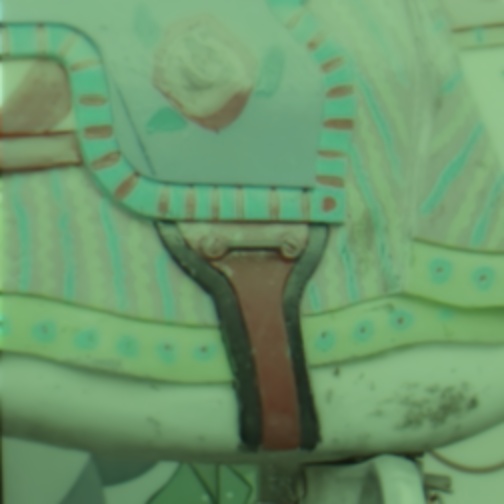}
                \includegraphics[width=\textwidth]{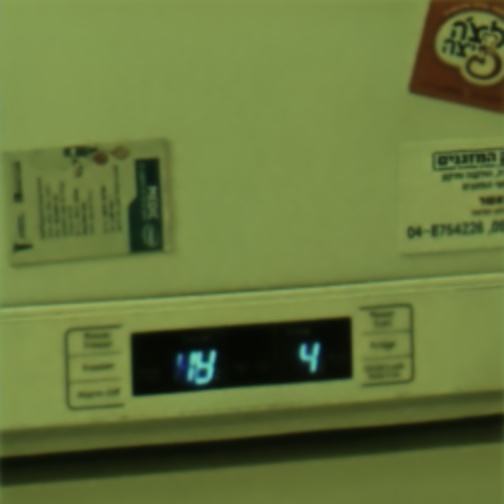}
                \includegraphics[width=\textwidth]{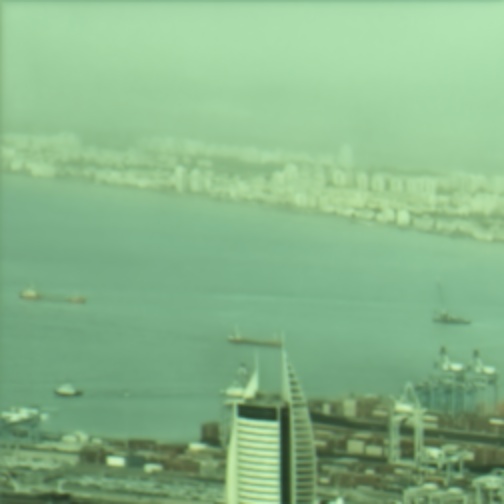}
                \label{fig:mi}
                \caption{MiAlgo}
        \end{subfigure}
        \begin{subfigure}[b]{0.11\textwidth}
               \includegraphics[width=\textwidth]{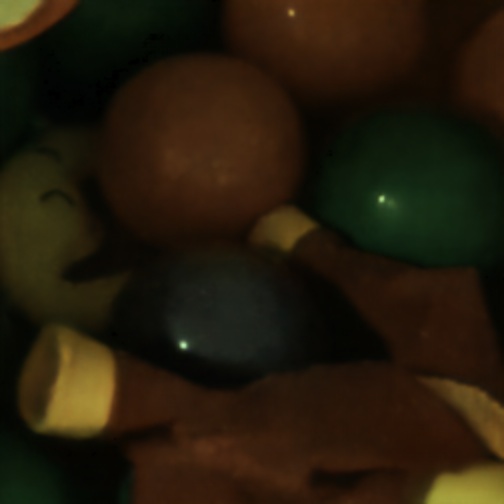}
                \includegraphics[width=\textwidth]{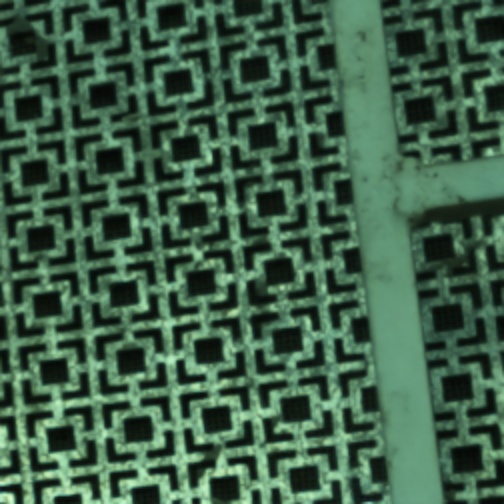}
                \includegraphics[width=\textwidth]{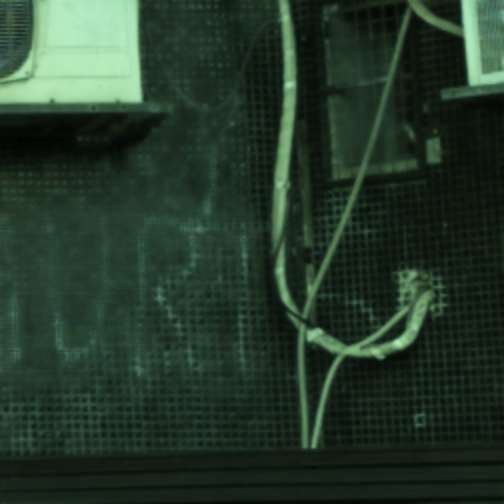}
                \includegraphics[width=\textwidth]{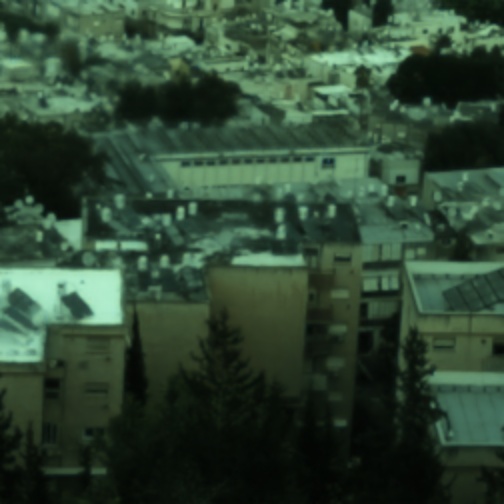}
                \includegraphics[width=\textwidth]{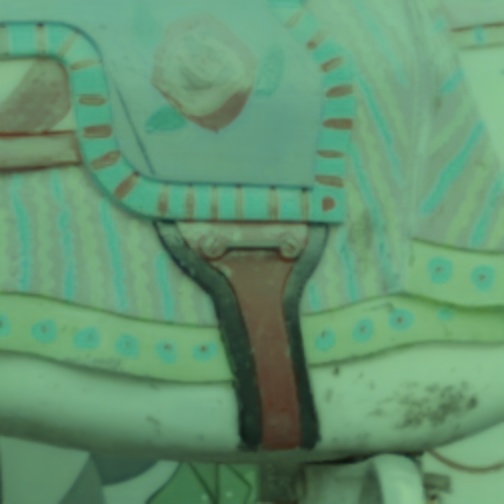}
                \includegraphics[width=\textwidth]{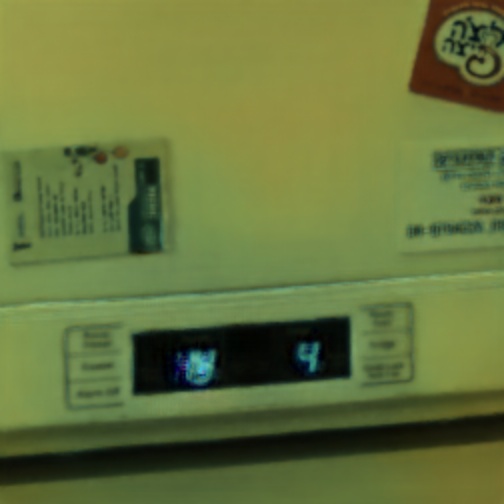}
                \includegraphics[width=\textwidth]{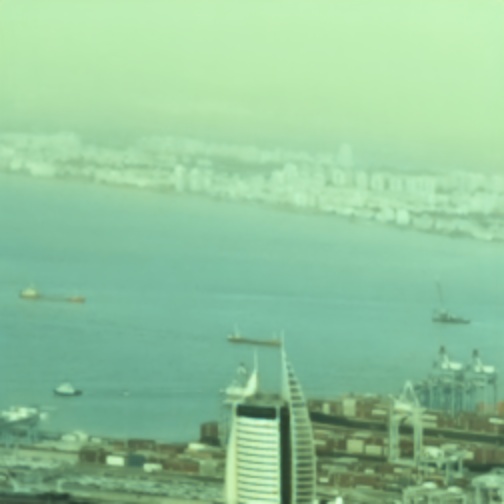}
                \label{fig:hiimage}
                \caption{HiImage}
        \end{subfigure}
        \begin{subfigure}[b]{0.11\textwidth}
                \includegraphics[width=\textwidth]{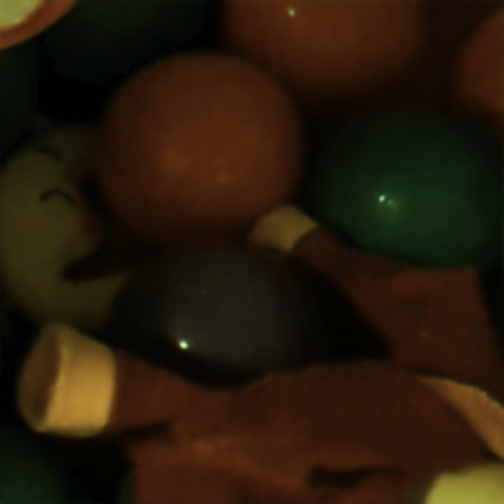}
                \includegraphics[width=\textwidth]{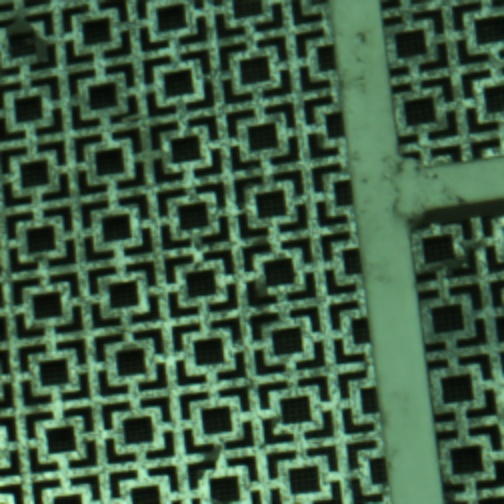}
                \includegraphics[width=\textwidth]{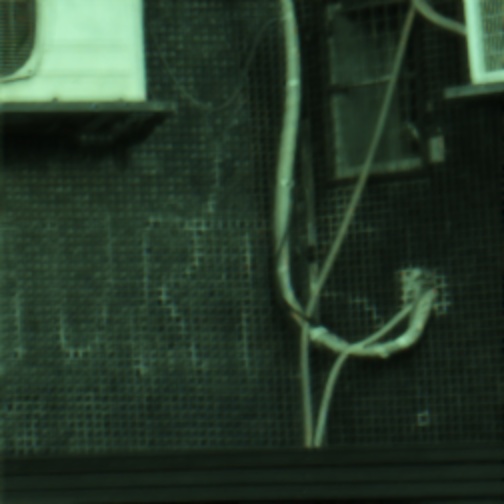}
                \includegraphics[width=\textwidth]{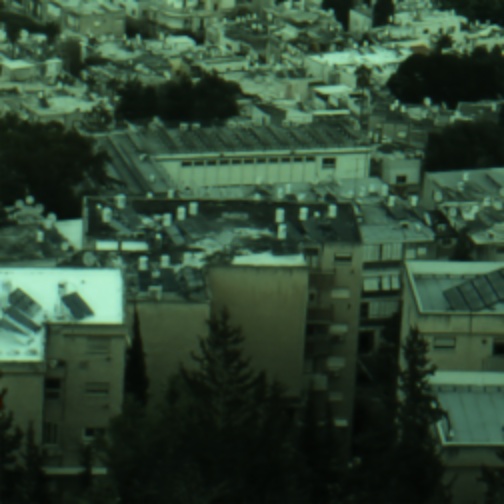}
                \includegraphics[width=\textwidth]{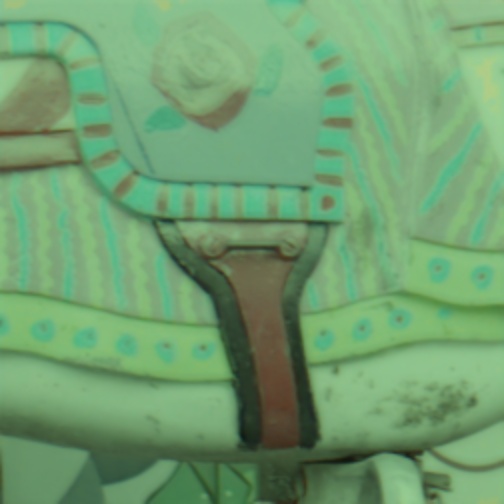}
                \includegraphics[width=\textwidth]{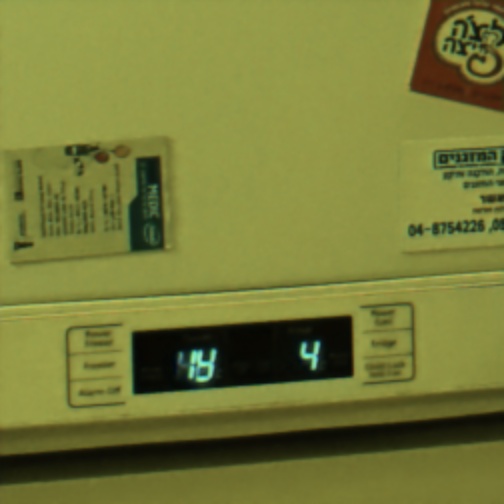}
                \includegraphics[width=\textwidth]{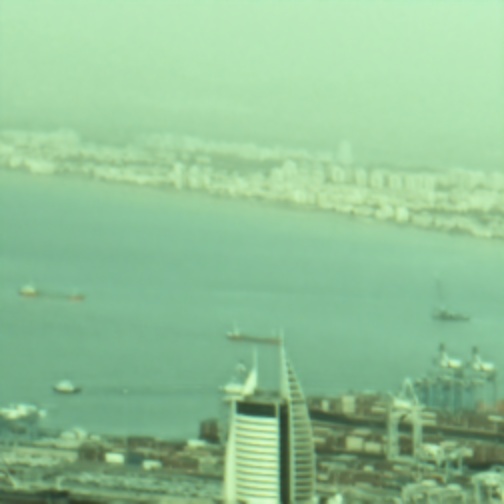}
                \label{fig:noah}
                \caption{NOAH}
        \end{subfigure}
        \begin{subfigure}[b]{0.11\textwidth}
                \includegraphics[width=\textwidth]{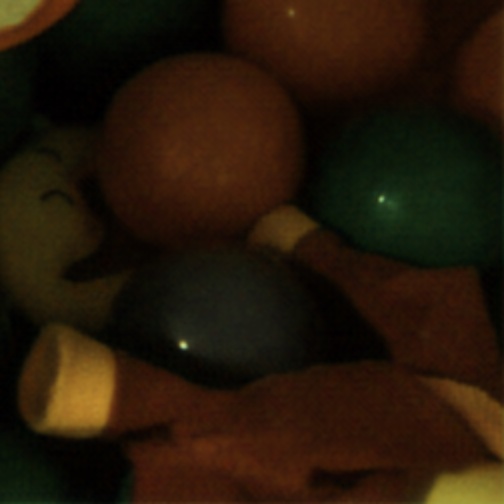}
                \includegraphics[width=\textwidth]{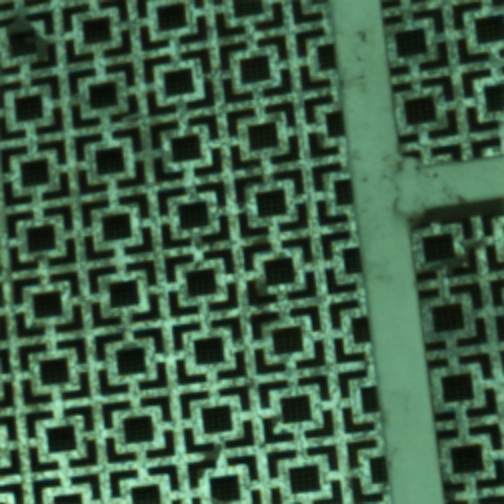}
                \includegraphics[width=\textwidth]{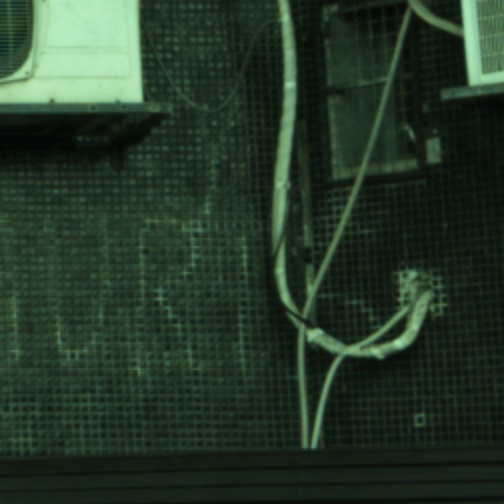}
                \includegraphics[width=\textwidth]{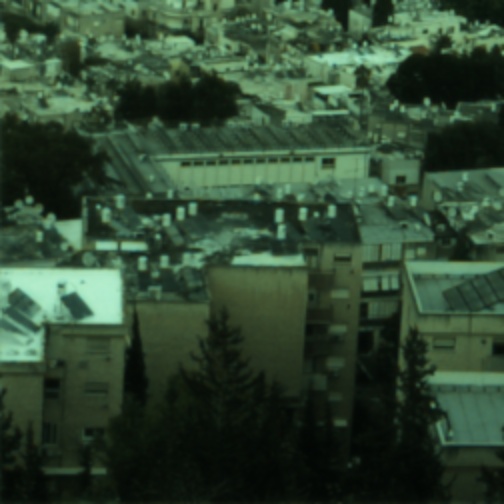}
                \includegraphics[width=\textwidth]{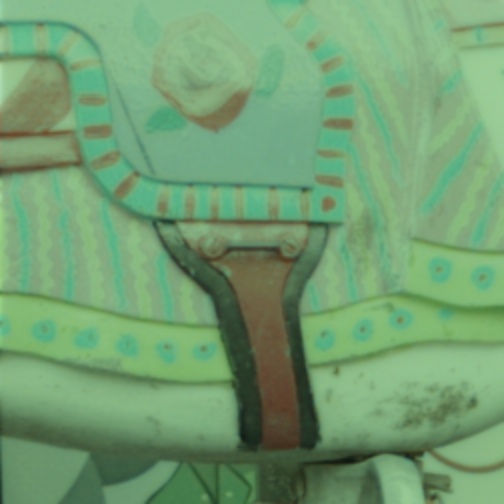}
                \includegraphics[width=\textwidth]{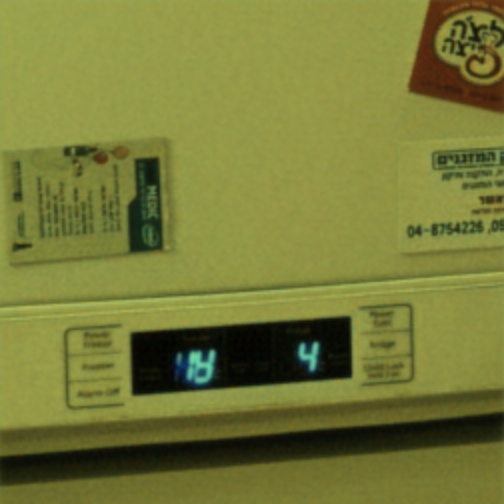}
                \includegraphics[width=\textwidth]{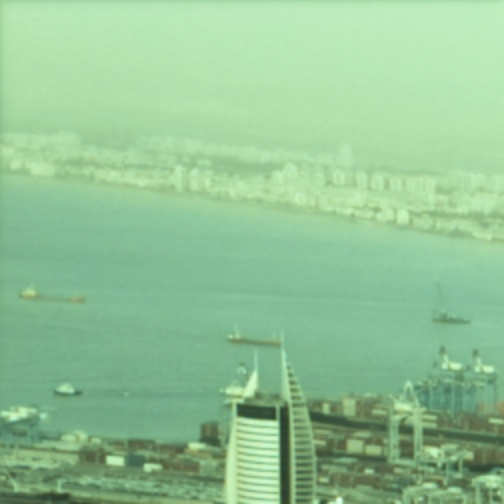}
                \label{fig:hit-iil}
                \caption{H-IIL}
        \end{subfigure}
        \begin{subfigure}[b]{0.11\textwidth}
                \includegraphics[width=\textwidth]{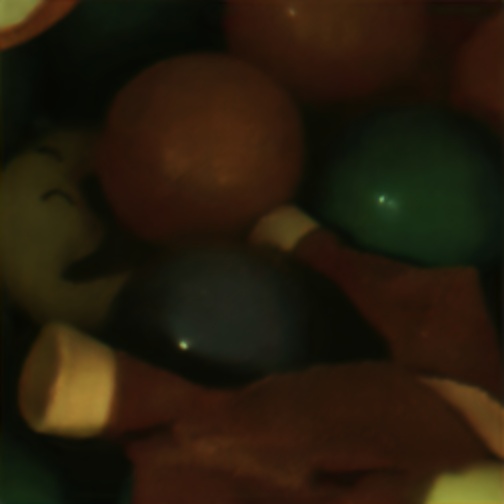}
                \includegraphics[width=\textwidth]{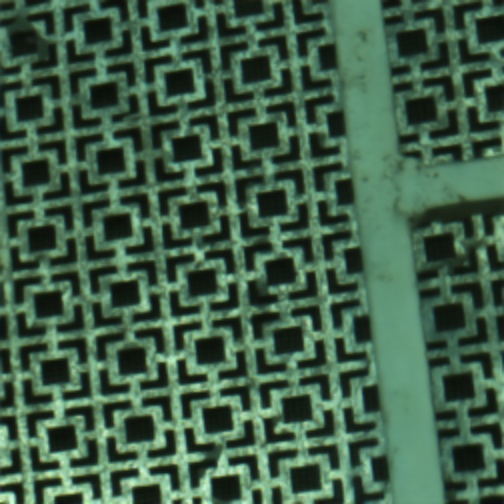}
                \includegraphics[width=\textwidth]{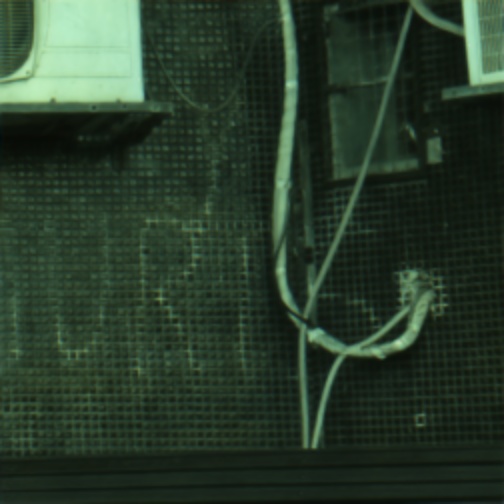}
                \includegraphics[width=\textwidth]{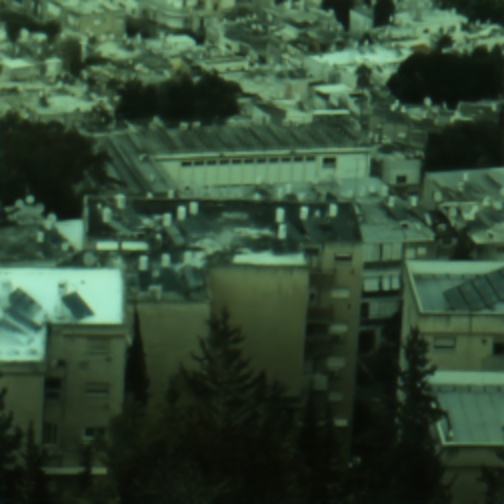}
                \includegraphics[width=\textwidth]{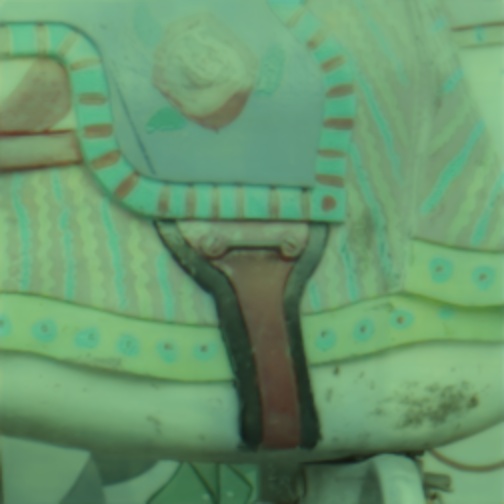}
                \includegraphics[width=\textwidth]{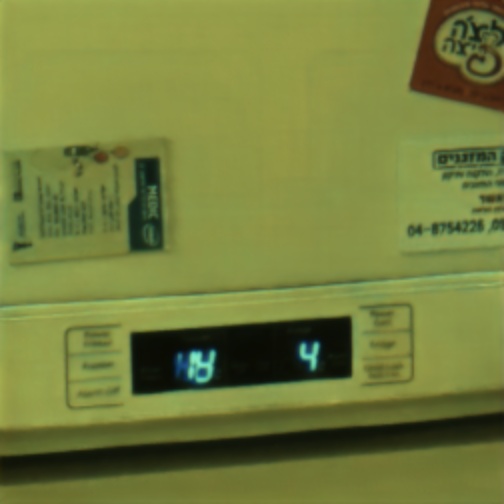}
                \includegraphics[width=\textwidth]{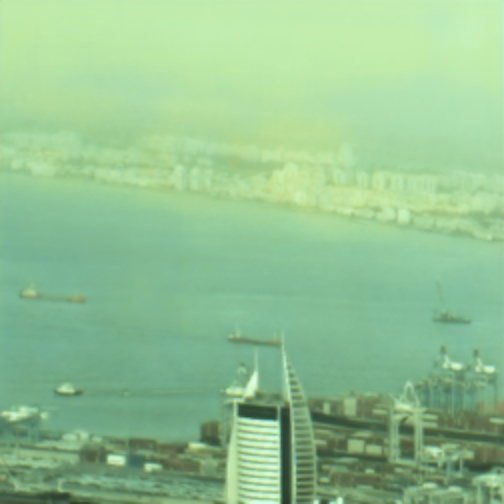}
                \label{fig:sense}
                \caption{Sense}
        \end{subfigure}
        \begin{subfigure}[b]{0.11\textwidth}
                \includegraphics[width=\textwidth]{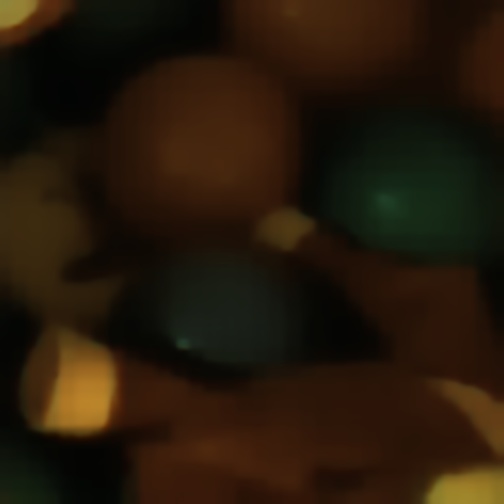}
                \includegraphics[width=\textwidth]{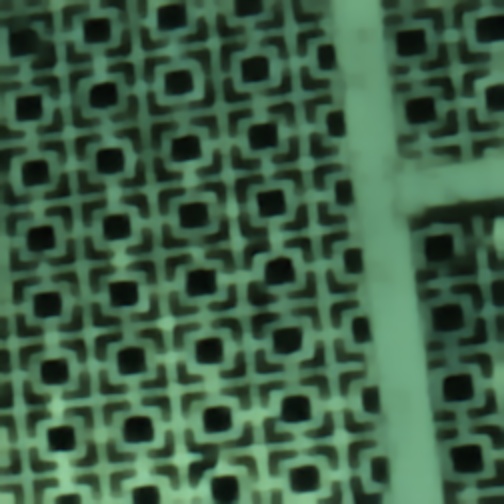}
                \includegraphics[width=\textwidth]{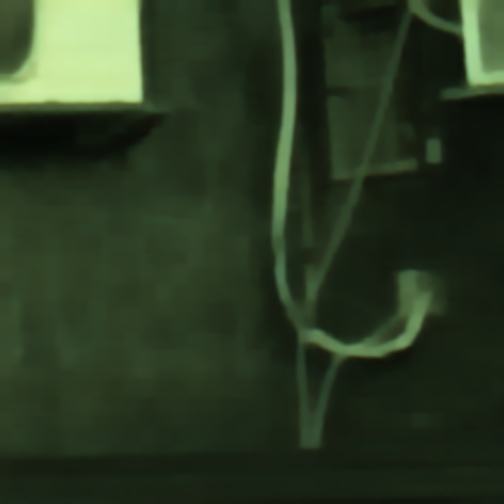}
                \includegraphics[width=\textwidth]{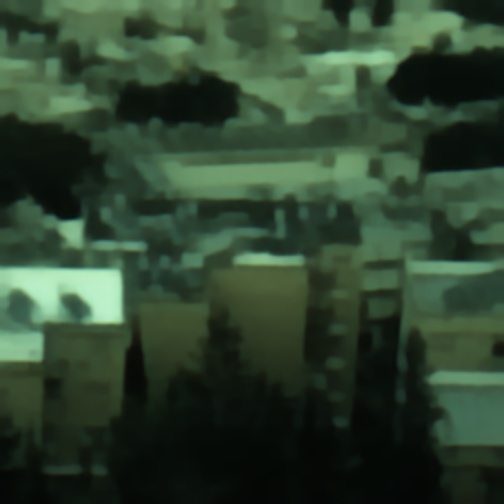}
                \includegraphics[width=\textwidth]{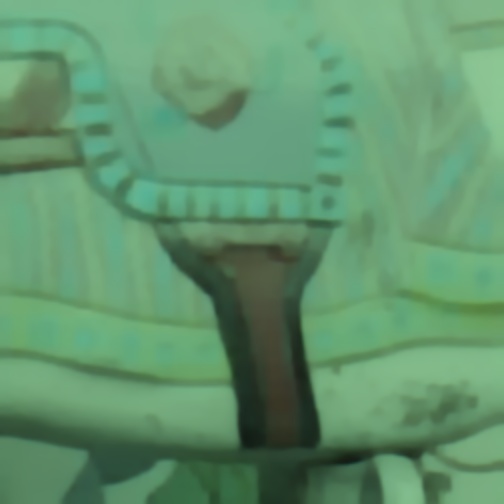}
                \includegraphics[width=\textwidth]{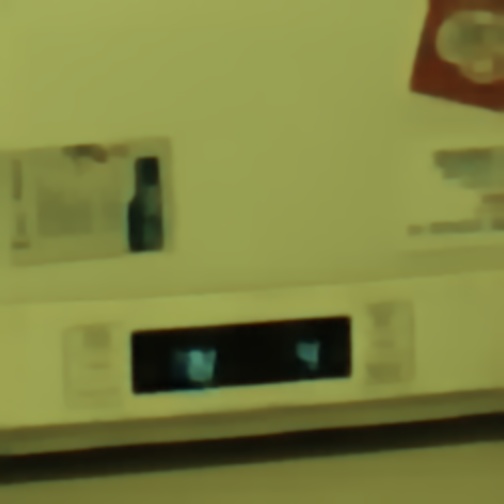}
                \includegraphics[width=\textwidth]{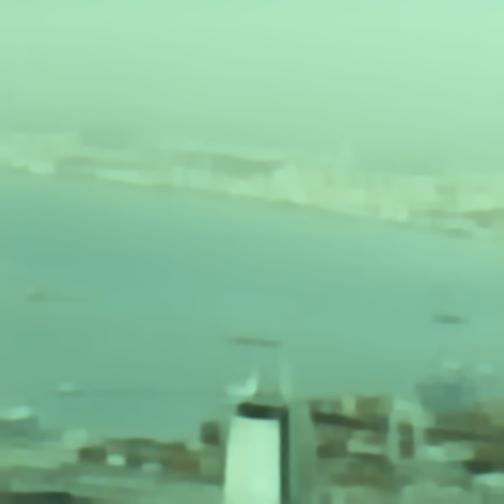}
                \label{fig:ozu}
                \caption{OzU}
        \end{subfigure}
         \caption{Qualitative results on Samsung S7 dataset, submitted by the participants.}
         \label{fig:qual-s7} 
\end{figure}

\subsection{Quantitative Results}

Following the evaluation methods given in AIM Reversed ISP challenge \cite{conde2022aim}, we have employed two common image similarity metrics in our experiments, which are namely Structural Similarity Index (SSIM) and Peak Signal-to-Noise Ratio (PSNR). Table \ref{tab:ablation} summarizes the performances of our proposed architecture and the other compared methods participating in the challenge. Although our results are not competitive in the rankings, we believe that investigating the idea of modeling the disruptive or modifying factors as the style factor [1] is an important and open-to-improvement approach. Thanks to the AIM Reversed ISP challenge \cite{conde2022aim}, we have had a chance to try our idea for this problem, which actually works seamlessly for the other domains, and to see what may be missing in our approach or which part of this idea could be improved. Note that we did not use any ensembling technique to boost the performance or any extra data in our experiments.

\captionsetup[subfigure]{font=small, labelformat=empty}
\begin{figure}[!t]
        \centering
        \begin{subfigure}[b]{0.11\textwidth}
                \includegraphics[width=\textwidth]{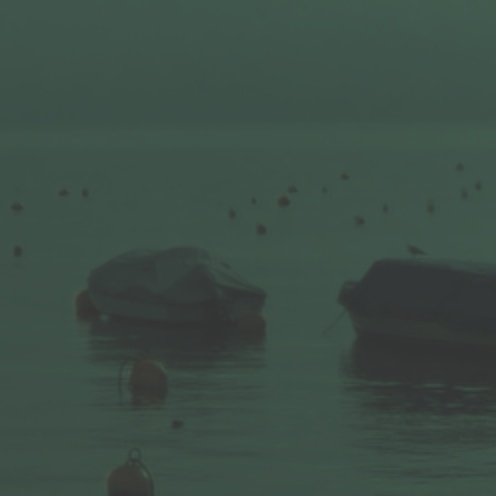}
                \includegraphics[width=\textwidth]{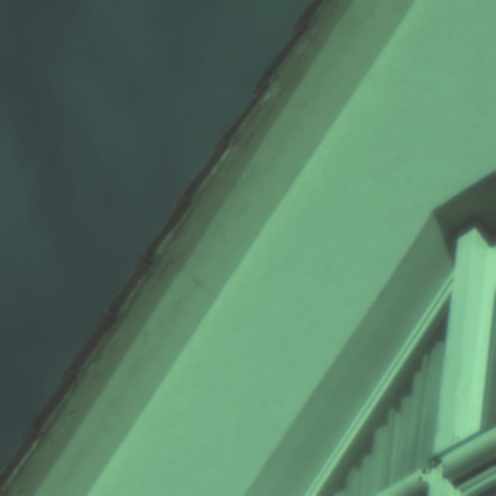}
                \includegraphics[width=\textwidth]{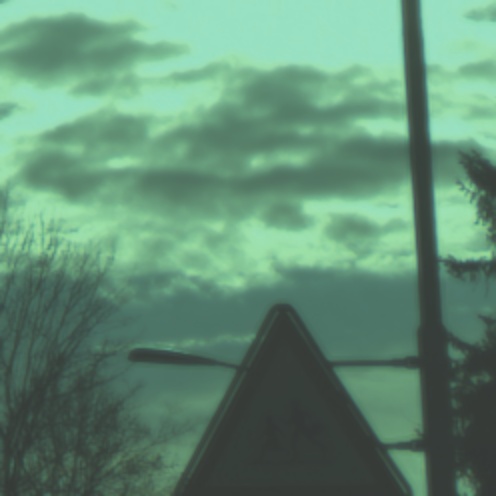}
                \includegraphics[width=\textwidth]{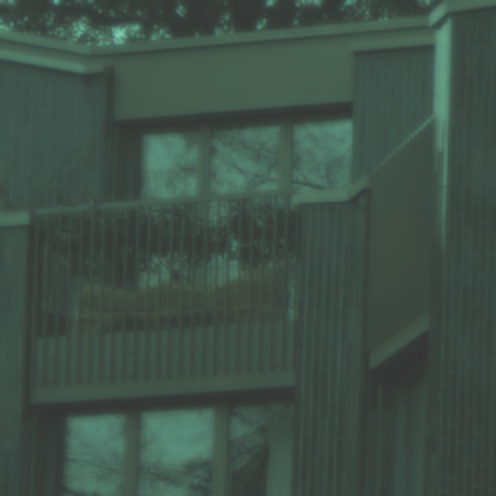}
                \includegraphics[width=\textwidth]{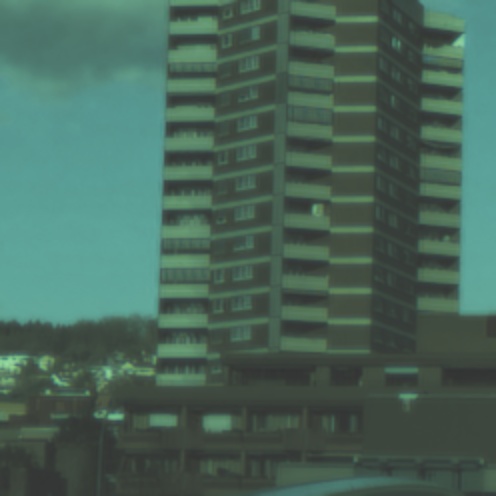}
                \includegraphics[width=\textwidth]{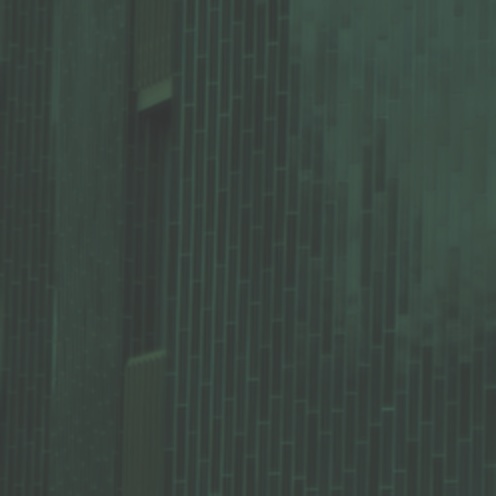}
                \includegraphics[width=\textwidth]{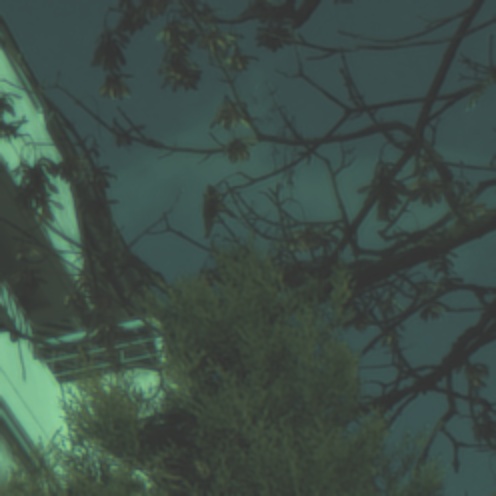}
                \includegraphics[width=\textwidth]{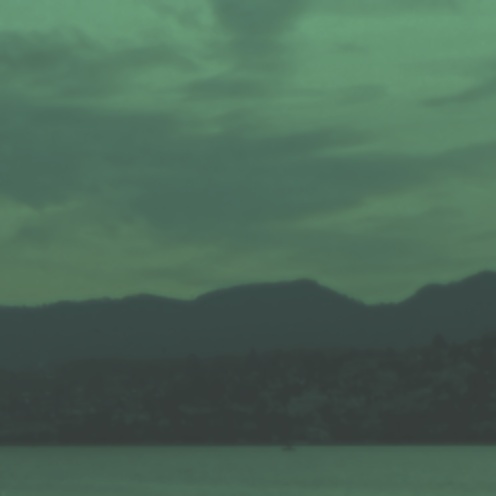}
                \label{fig:0noise-qual}
                \caption{0noise}
        \end{subfigure}  
        \begin{subfigure}[b]{0.11\textwidth}
                \includegraphics[width=\textwidth]{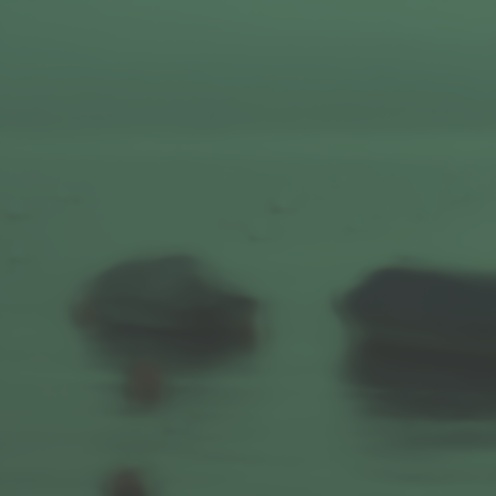}
                \includegraphics[width=\textwidth]{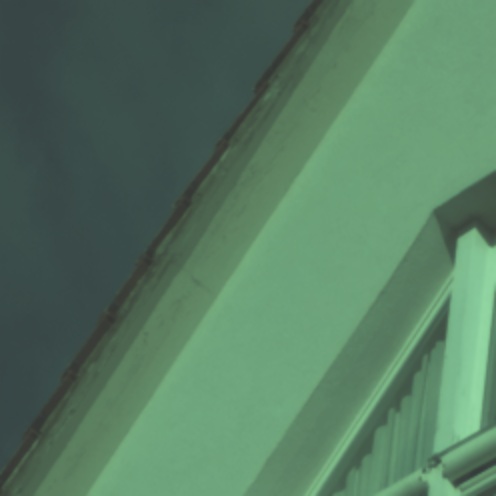}
                \includegraphics[width=\textwidth]{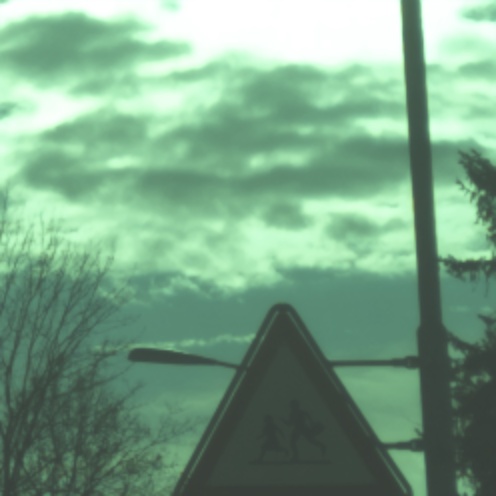}
                \includegraphics[width=\textwidth]{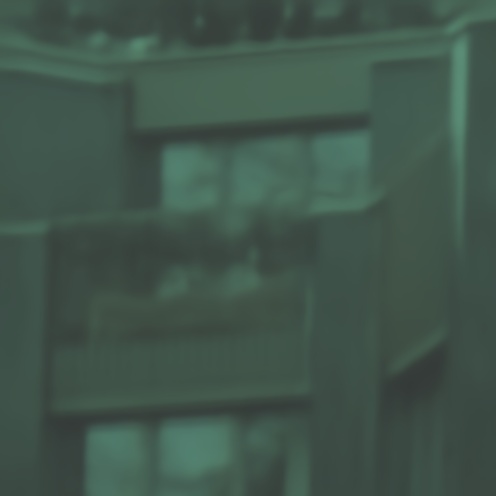}
                \includegraphics[width=\textwidth]{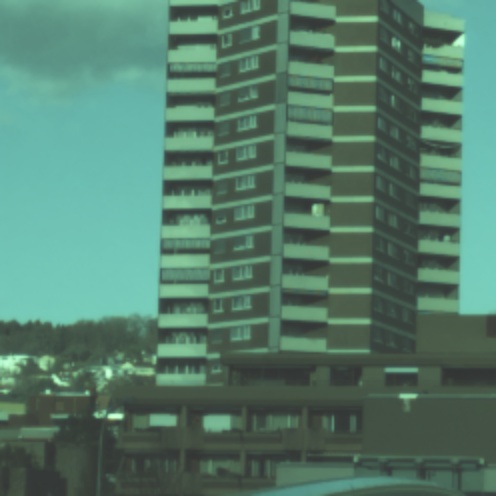}
                \includegraphics[width=\textwidth]{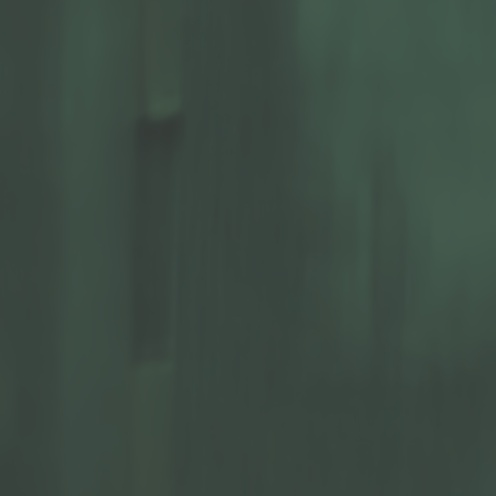}
                \includegraphics[width=\textwidth]{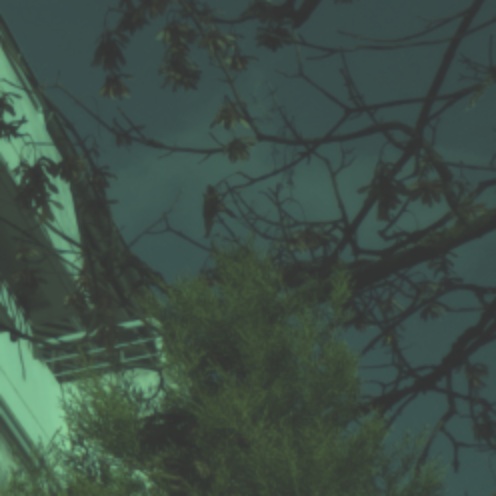}
                \includegraphics[width=\textwidth]{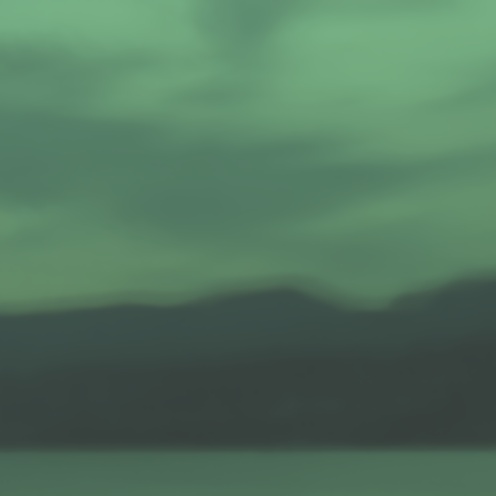}
                \label{fig:casia-qual}
                \caption{CASIA}
        \end{subfigure}       
        \begin{subfigure}[b]{0.11\textwidth}
              \includegraphics[width=\textwidth]{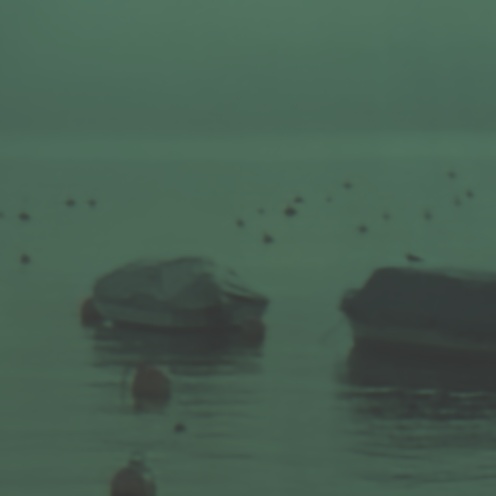}
                \includegraphics[width=\textwidth]{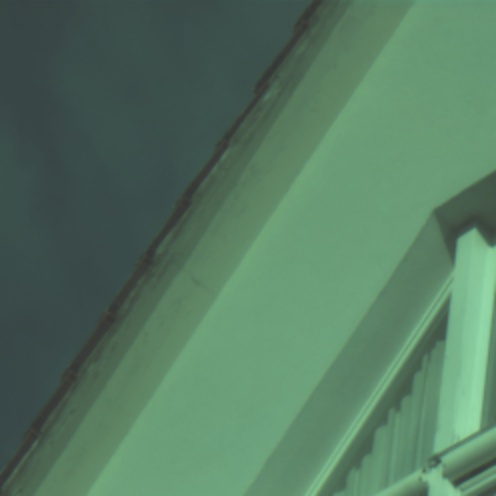}
                \includegraphics[width=\textwidth]{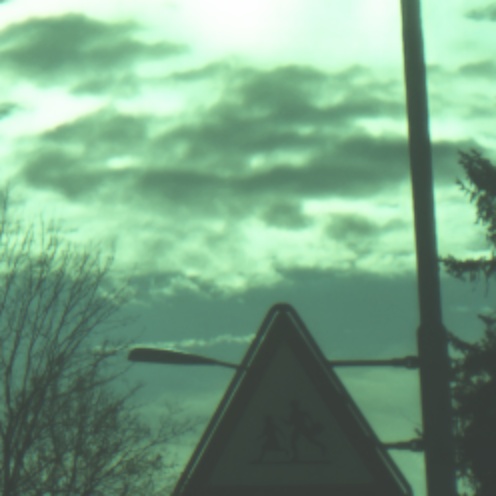}
                \includegraphics[width=\textwidth]{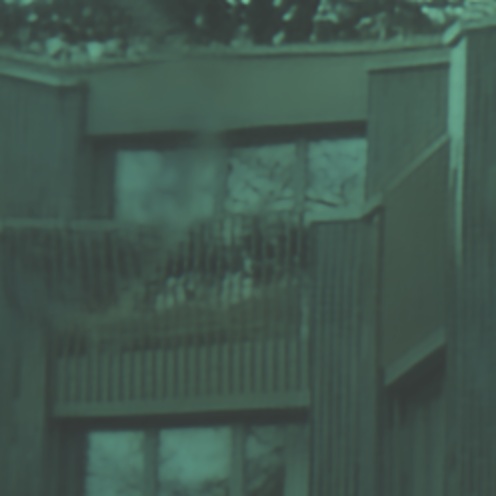}
                \includegraphics[width=\textwidth]{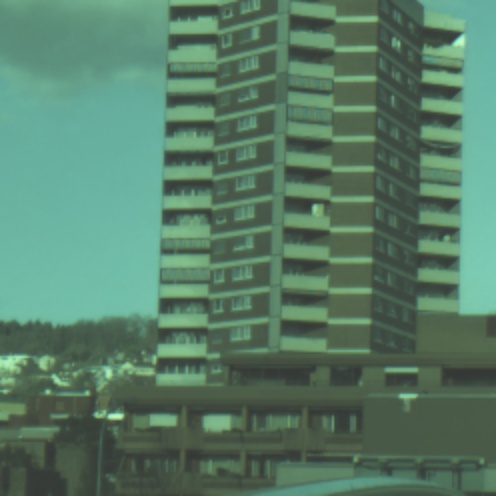}
                \includegraphics[width=\textwidth]{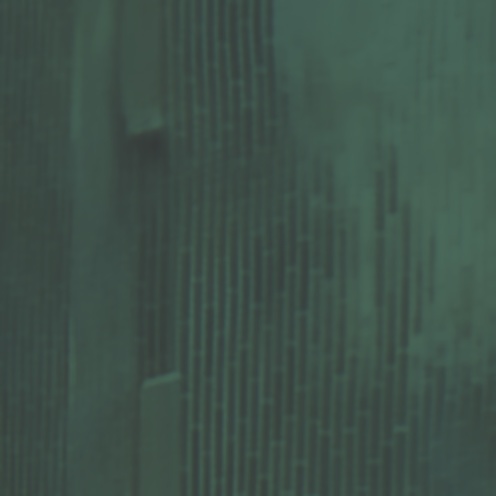}
                \includegraphics[width=\textwidth]{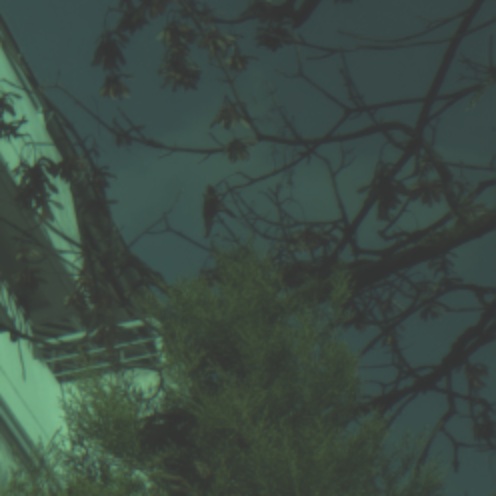}
                \includegraphics[width=\textwidth]{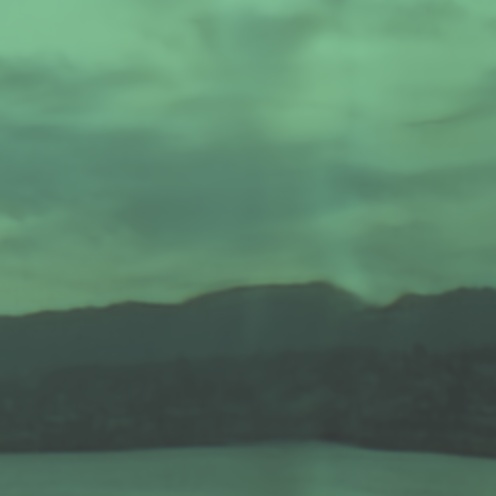}
                \label{fig:mi-qual}
                \caption{MiAlgo}
        \end{subfigure}
        \begin{subfigure}[b]{0.11\textwidth}
               \includegraphics[width=\textwidth]{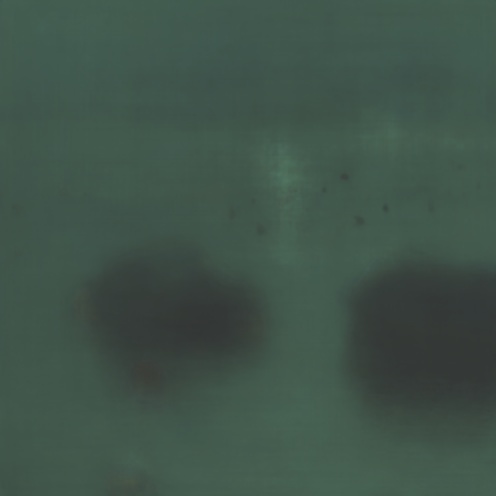}
                \includegraphics[width=\textwidth]{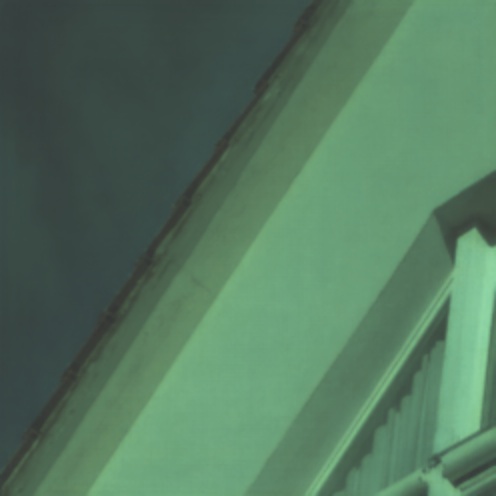}
                \includegraphics[width=\textwidth]{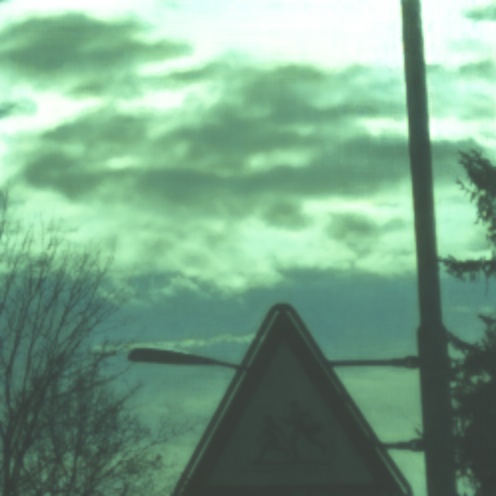}
                \includegraphics[width=\textwidth]{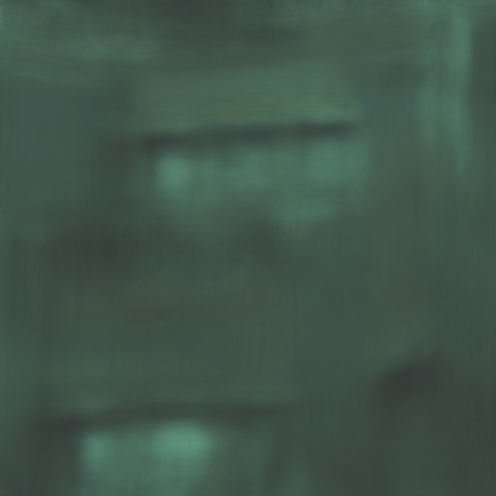}
                \includegraphics[width=\textwidth]{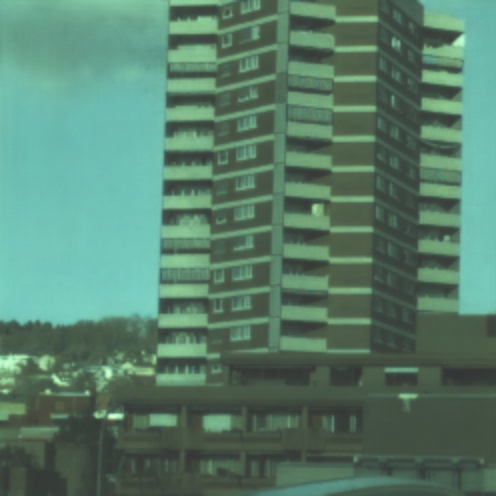}
                \includegraphics[width=\textwidth]{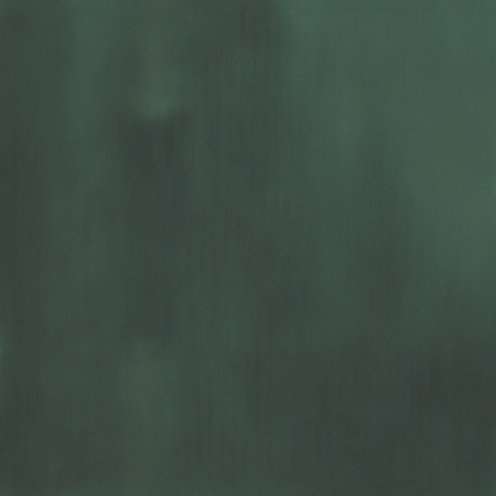}
                \includegraphics[width=\textwidth]{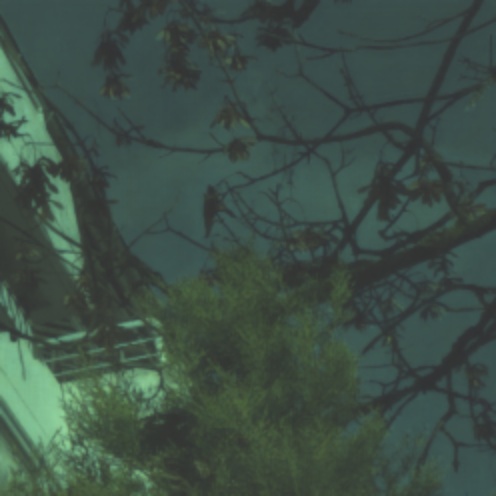}
                \includegraphics[width=\textwidth]{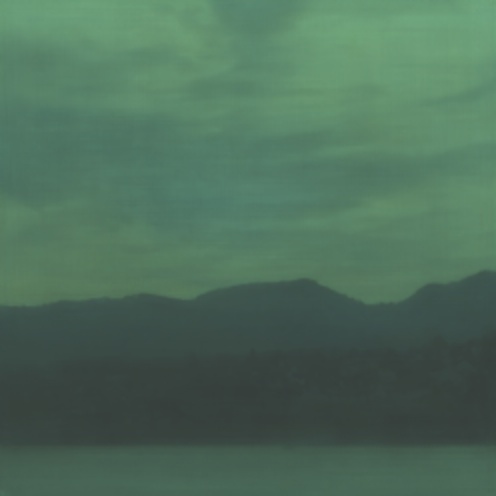}
                \label{fig:hiimage-qual}
                \caption{HiImage}
        \end{subfigure}
        \begin{subfigure}[b]{0.11\textwidth}
                \includegraphics[width=\textwidth]{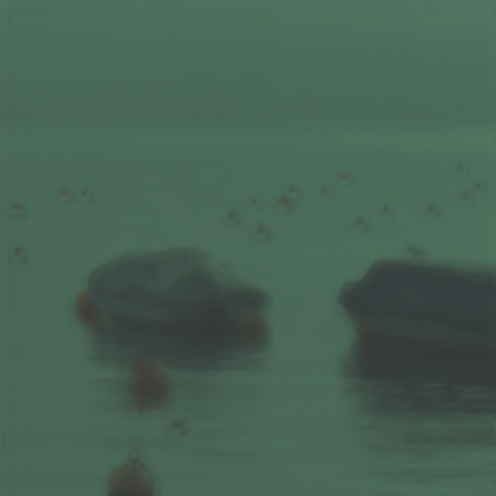}
                \includegraphics[width=\textwidth]{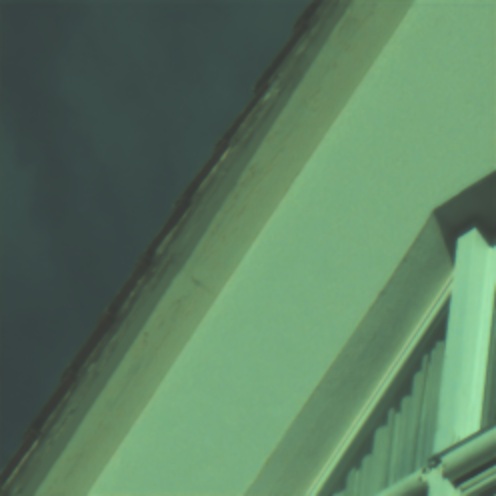}
                \includegraphics[width=\textwidth]{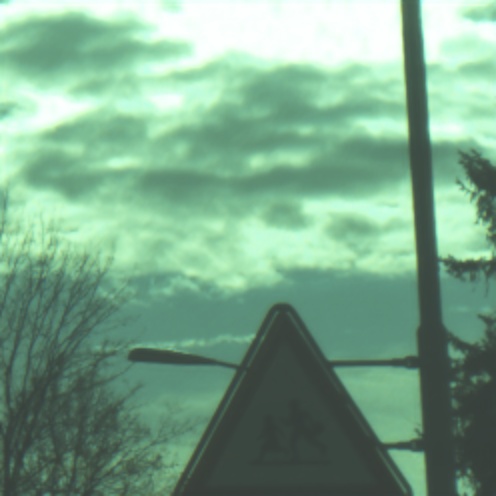}
                \includegraphics[width=\textwidth]{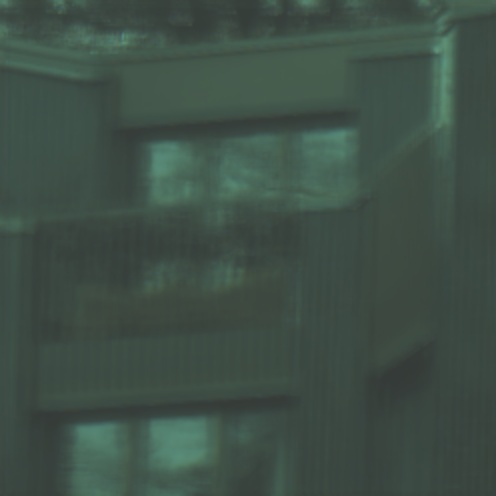}
                \includegraphics[width=\textwidth]{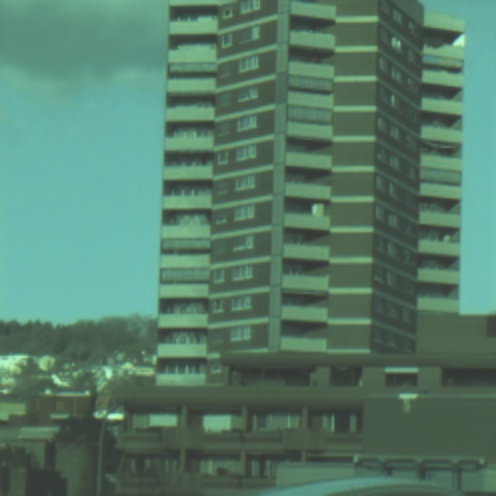}
                \includegraphics[width=\textwidth]{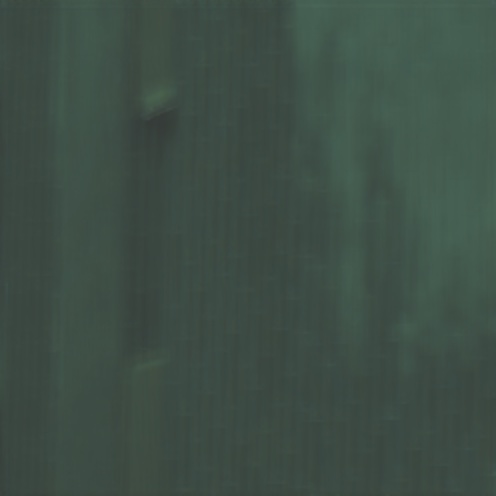}
                \includegraphics[width=\textwidth]{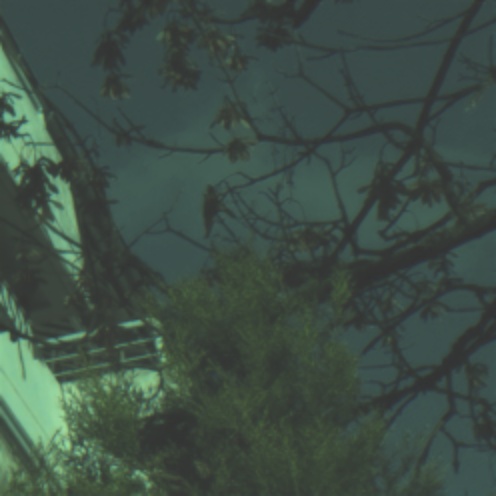}
                \includegraphics[width=\textwidth]{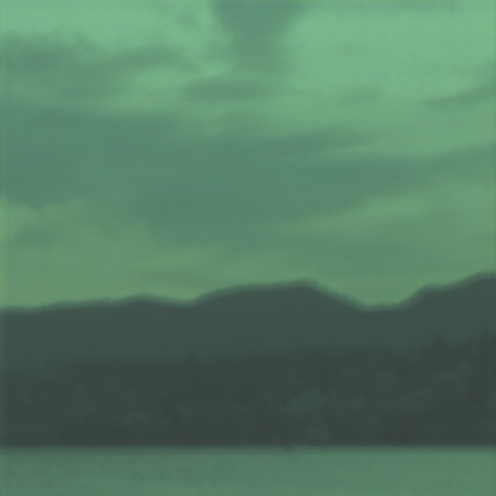}
                \label{fig:noah-qual}
                \caption{NOAH}
        \end{subfigure}
        \begin{subfigure}[b]{0.11\textwidth}
                \includegraphics[width=\textwidth]{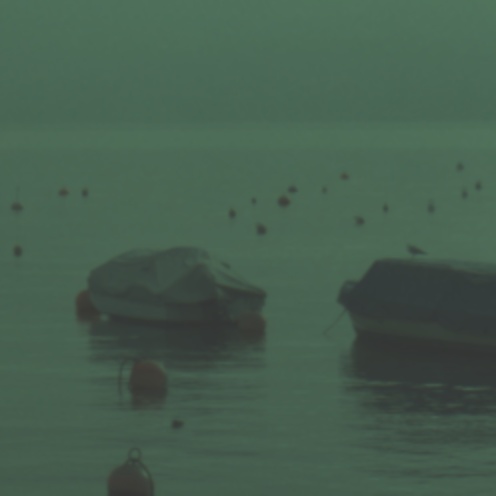}
                \includegraphics[width=\textwidth]{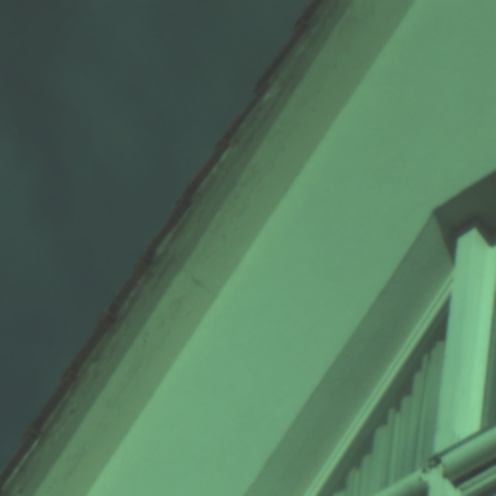}
                \includegraphics[width=\textwidth]{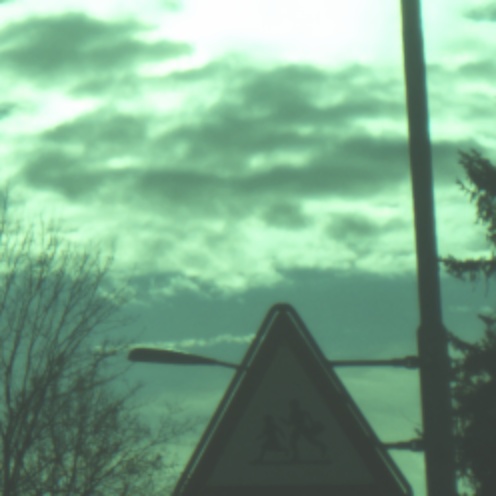}
                \includegraphics[width=\textwidth]{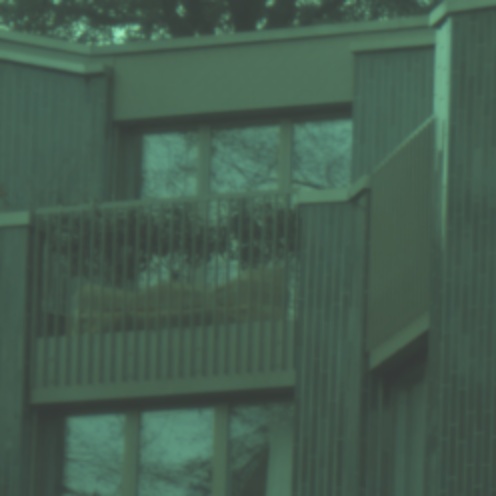}
                \includegraphics[width=\textwidth]{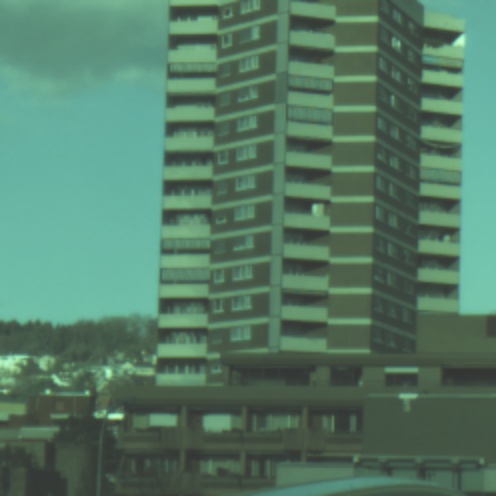}
                \includegraphics[width=\textwidth]{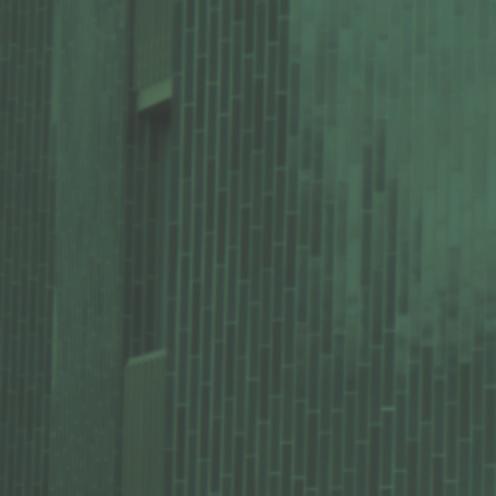}
                \includegraphics[width=\textwidth]{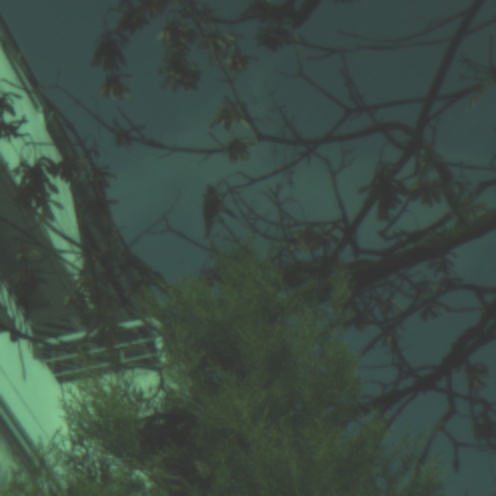}
                \includegraphics[width=\textwidth]{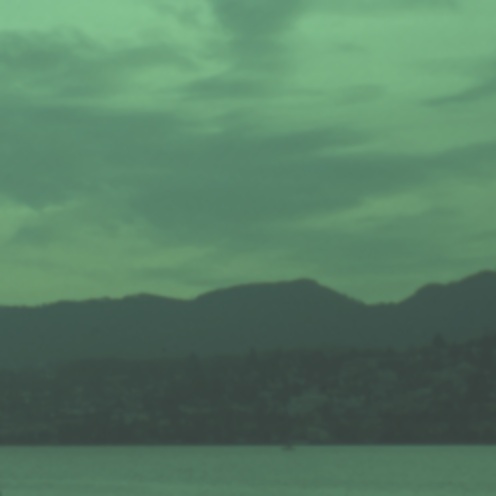}
                \label{fig:hit-iil-qual}
                \caption{H-IIL}
        \end{subfigure}
        \begin{subfigure}[b]{0.11\textwidth}
                \includegraphics[width=\textwidth]{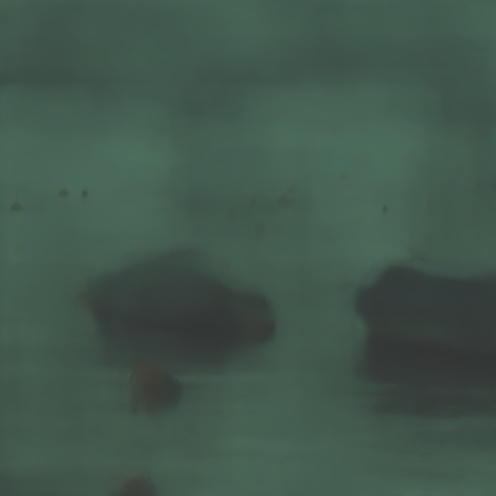}
                \includegraphics[width=\textwidth]{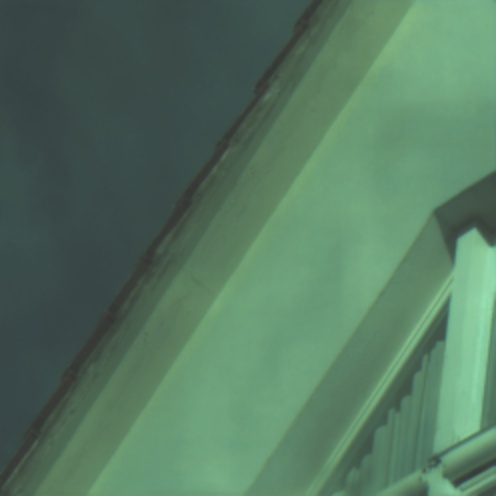}
                \includegraphics[width=\textwidth]{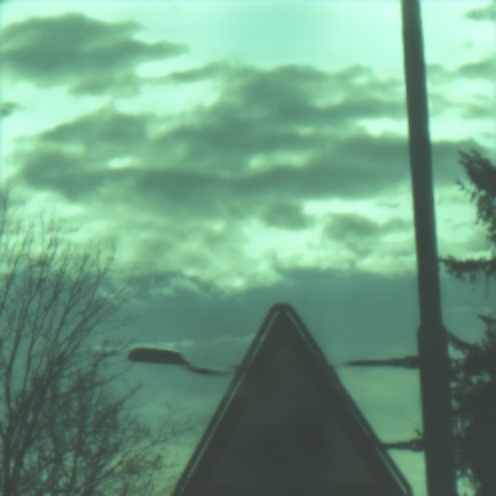}
                \includegraphics[width=\textwidth]{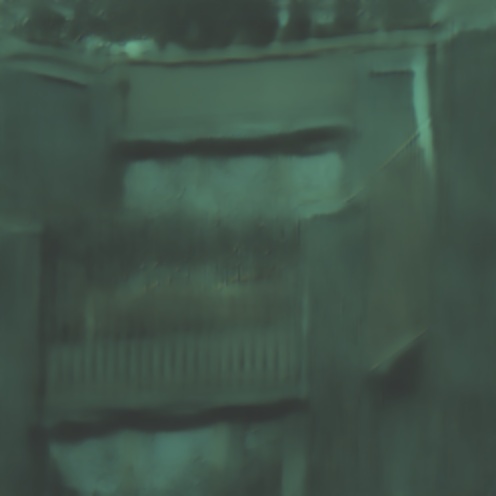}
                \includegraphics[width=\textwidth]{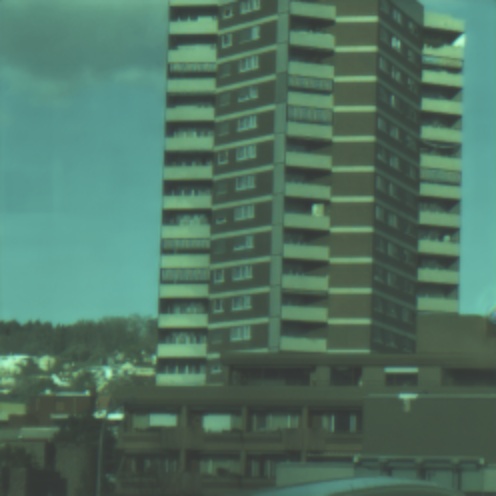}
                \includegraphics[width=\textwidth]{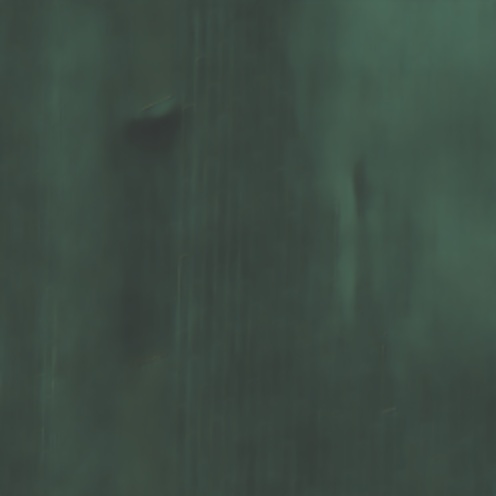}
                \includegraphics[width=\textwidth]{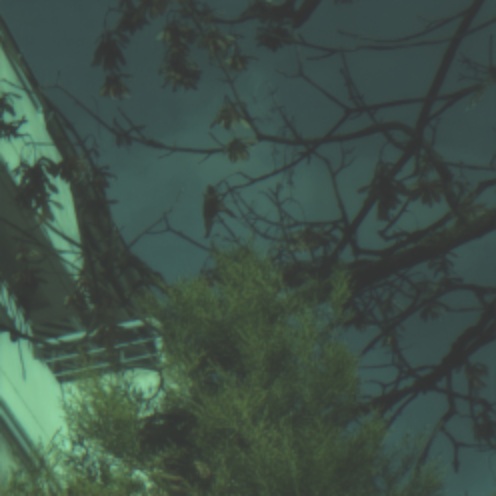}
                \includegraphics[width=\textwidth]{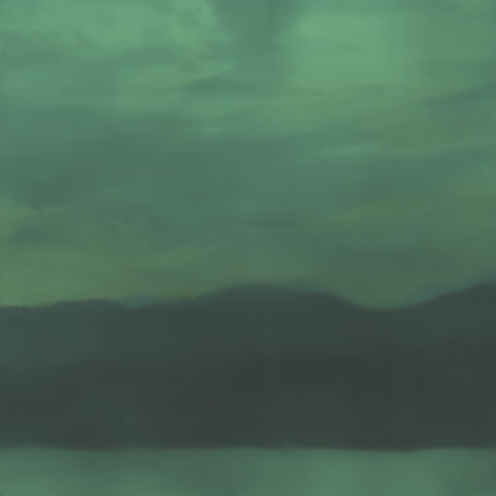}
                \label{fig:sense-qual}
                \caption{Sense}
        \end{subfigure}
        \begin{subfigure}[b]{0.11\textwidth}
                \includegraphics[width=\textwidth]{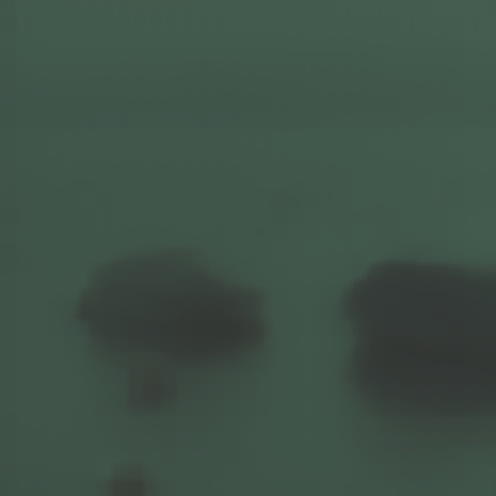}
                \includegraphics[width=\textwidth]{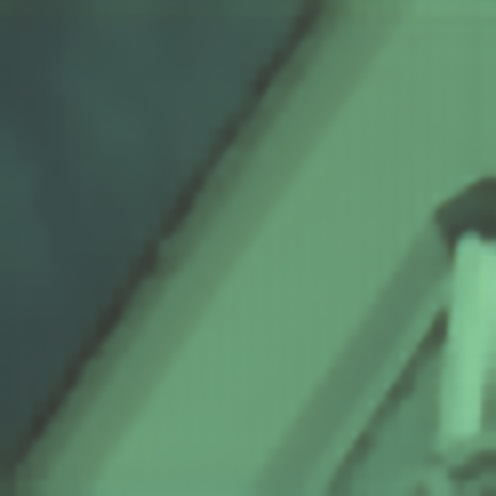}
                \includegraphics[width=\textwidth]{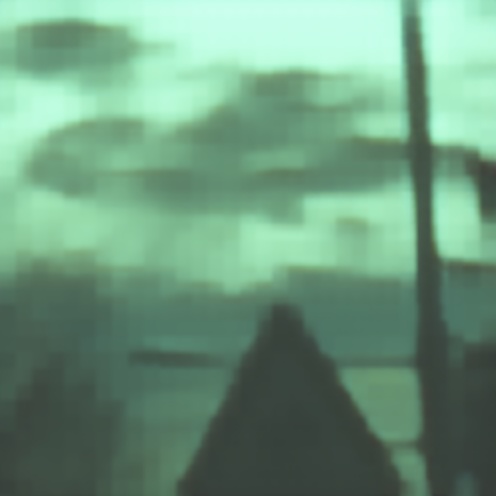}
                \includegraphics[width=\textwidth]{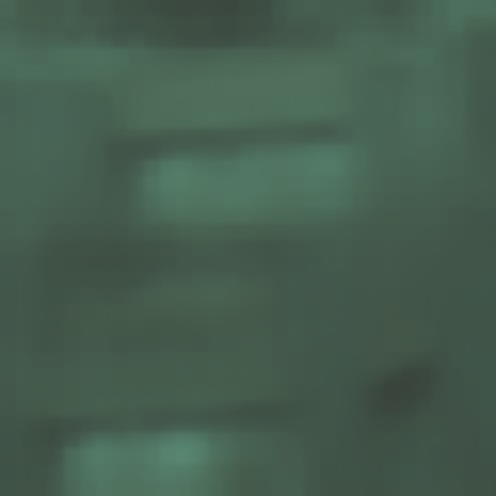}
                \includegraphics[width=\textwidth]{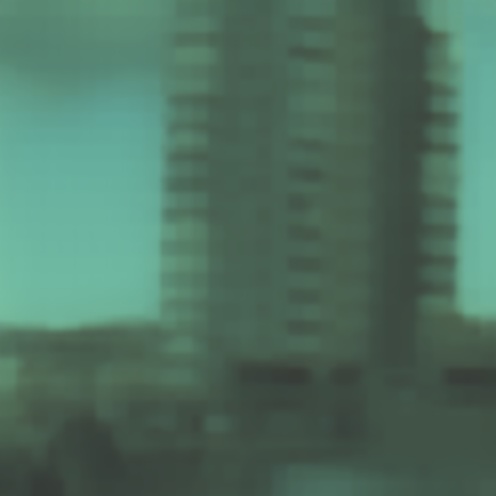}
                \includegraphics[width=\textwidth]{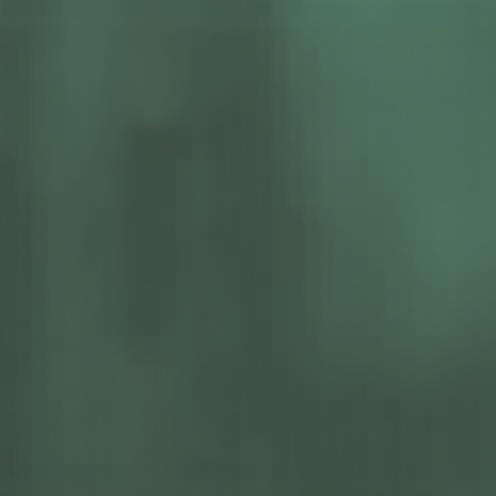}
                \includegraphics[width=\textwidth]{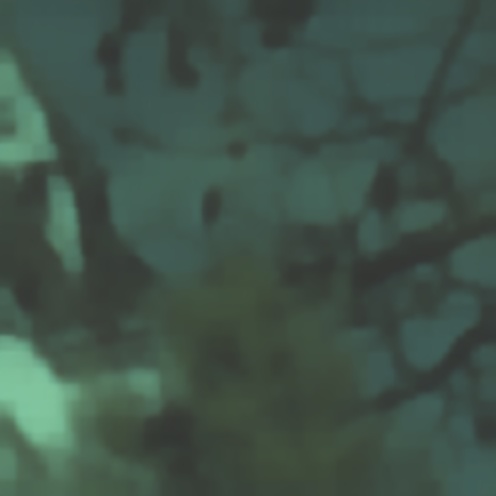}
                \includegraphics[width=\textwidth]{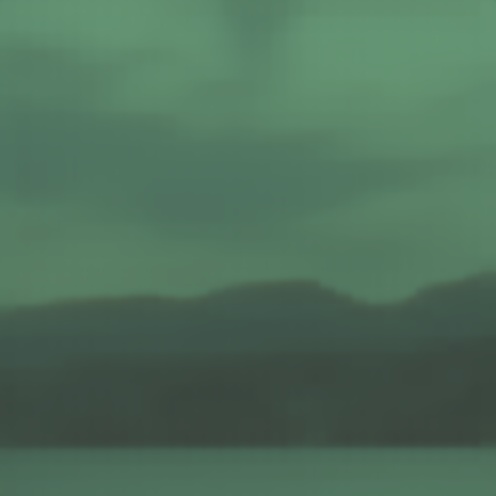}
                \label{fig:ozu-qual}
                \caption{OzU}
        \end{subfigure}
         \caption{Qualitative results on ETH Huawei P20 dataset, submitted by the participants.}
         \label{fig:qual-p20} 
\end{figure}

\subsection{Qualitative Comparison}

Figure \ref{fig:qual-s7} presents the qualitative results of the participants on the Samsung S7 dataset. Our proposed solution does not have superior performance on the task of reversing the ISP operations. However, our main goal in this study is to investigate how performing the idea of modeling the ISP operations as the style factor without using any ensembling strategy or extra data. In the previous studies \cite{Kinli_2021_CVPR,Kinli_2022_CVPR}, the idea of modeling any disruptive or modifying factor as the style factor works well without requiring handling the details. According to our observation, removing the effects of the ISP operations by adaptive feature normalization leads to losing the high-frequency details in the output. This significantly reduces the overall performance of our strategy for this task. At this point, we have tried to employ Wavelet-based discriminators for our adversarial training, yet it still gives out the performance. Moreover, following the previous studies on modeling as the style factor, we apply an adversarial training strategy for our experiments to achieve discriminative regularization on the output. As we notice in these results, we have to reconsider to use of this strategy for such a task since, to the best of our knowledge, the leading participants do not use this kind of strategy in their studies, and also it makes the training process more complicated for convergence. Nevertheless, we believe that the style of being sRGB is successfully reverted back to the style of being RAW in our final outputs, and this shows that the idea is still valid, but needs to be improved while building the architecture.

As shown in Figure \ref{fig:qual-p20}, similar problems on ETH Huawei P20 dataset come to the forefront. The finding that can be significant for our future studies is that the output is more likely to be the strongly-blurred version of the ground-truth RAW image. Although the style injected into the RAW image (\textit{i.e.}, the effect of multiple non-linear operations applied to the RAW image) is more or less successfully modeled and removed to translate the sRGB image to its RAW version, the blurring effect mostly leads to losing the high-frequency details in the output, so the
decrease in the performance for challenge benchmark. This effect is far stronger than the one that we face for the Track S7. We think that the alignment issue among the pairs for this dataset and not using any specialized module that resolves this issue may lead to amplifying the problem in our strategy. Another reason for the amplification of this issue could be to use an adversarial training strategy since the real and fake images suffer from being the exactly same for the discriminator and it does not help the training at all, even making it worse.

\section{Conclusion}
\label{sec:concl}

In this study, we investigate the performance of the approach of modeling the ISP operations as the style factor. We have followed the previous strategies to model the disruptive or modifying factors (\textit{i.e.}, multiple non-linear transformations applied during the ISP). Experiments on Samsung S7 and ETH Huawei P20 datasets show that this approach requires to be improved to be more competitive among the recent studies presented in the AIM Reversed ISP challenge \cite{conde2022aim}. In spite of the lack of performance in this task, compared to the recent methods in this challenge, the results indicate that the idea of modeling the disruptive factors as the style factor is still promising. Also, this challenge gives us a chance to see the weaknesses of our approach and encourages us to re-think this idea. The main limitation of our approach is the higher number of parameters composing the architecture to learn the style factor, and this can be reduced by using different feature distribution representations for the style factor. We will re-consider the usage of discriminative regularization for such tasks. For future work, we will re-design the whole architecture, the way of expressing the style information, and the objective function used and continue to investigate the style factor in reconstruction tasks.

%
%
\bibliographystyle{splncs04}
\bibliography{egbib}

\begin{thebibliography}{10}
\providecommand{\url}[1]{\texttt{#1}}
\providecommand{\urlprefix}{URL }
\providecommand{\doi}[1]{https://doi.org/#1}

\bibitem{brooks2019unprocessing}
Brooks, T., Mildenhall, B., Xue, T., Chen, J., Sharlet, D., Barron, J.T.:
  Unprocessing images for learned raw denoising. In: Proceedings of the
  IEEE/CVF Conference on Computer Vision and Pattern Recognition. pp.
  11036--11045 (2019)

\bibitem{conde2022model}
Conde, M.V., McDonagh, S., Maggioni, M., Leonardis, A., P{\'e}rez-Pellitero,
  E.: Model-based image signal processors via learnable dictionaries. In:
  Proceedings of the AAAI Conference on Artificial Intelligence. vol.~36, pp.
  481--489 (2022)

\bibitem{conde2022aim}
Conde, M.V., Timofte, R., et~al.: Reversed image signal processing and raw
  reconstruction. aim 2022 challenge report. In: Proceedings of the European
  Conference on Computer Vision Workshops (ECCVW) (2022)

\bibitem{deng2009imagenet}
Deng, J., Dong, W., Socher, R., Li, L.J., Li, K., Fei-Fei, L.: Imagenet: A
  large-scale hierarchical image database. In: 2009 IEEE conference on computer
  vision and pattern recognition. pp. 248--255. Ieee (2009)

\bibitem{Gatys2015c}
Gatys, L.A., Ecker, A.S., Bethge, M.: A neural algorithm of artistic style.
  arXiv  (Aug 2015), \url{http://arxiv.org/abs/1508.06576}

\bibitem{46163}
Ghiasi, G., Lee, H., Kudlur, M., Dumoulin, V., Shlens, J.: Exploring the
  structure of a real-time, arbitrary neural artistic stylization network
  (2017), \url{https://arxiv.org/abs/1705.06830}

\bibitem{huang2017adain}
Huang, X., Belongie, S.: Arbitrary style transfer in real-time with adaptive
  instance normalization. In: ICCV (2017)

\bibitem{ignatov2022isp}
Ignatov, A., Timofte, R., et~al.: Learned smartphone isp on mobile gpus with
  deep learning, mobile ai \& aim 2022 challenge: Report. In: Proceedings of
  the European Conference on Computer Vision (ECCV) Workshops (2022)

\bibitem{ignatov2020replacing}
Ignatov, A., Van~Gool, L., Timofte, R.: Replacing mobile camera isp with a
  single deep learning model. In: Proceedings of the IEEE/CVF Conference on
  Computer Vision and Pattern Recognition Workshops. pp. 536--537 (2020)

\bibitem{DBLP:journals/corr/KingmaB14}
Kingma, D.P., Ba, J.: Adam: {A} method for stochastic optimization. In: Bengio,
  Y., LeCun, Y. (eds.) 3rd International Conference on Learning
  Representations, {ICLR} 2015, San Diego, CA, USA, May 7-9, 2015, Conference
  Track Proceedings (2015), \url{http://arxiv.org/abs/1412.6980}

\bibitem{Kinli_2022_CVPR}
K{\i}nl{\i}, F., \"Ozcan, B., K{\i}ra\c{c}, F.: Patch-wise contrastive style
  learning for instagram filter removal. In: Proceedings of the IEEE/CVF
  Conference on Computer Vision and Pattern Recognition (CVPR) Workshops. pp.
  578--588 (June 2022)

\bibitem{Kinli_2021_CVPR}
Kinli, F., Ozcan, B., Kirac, F.: Instagram filter removal on fashionable
  images. In: Proceedings of the IEEE/CVF Conference on Computer Vision and
  Pattern Recognition (CVPR) Workshops. pp. 736--745 (June 2021)

\bibitem{lamb2016discriminative}
Lamb, A., Dumoulin, V., Courville, A.: Discriminative regularization for
  generative models. arXiv preprint arXiv:1602.03220  (2016)

\bibitem{lin2014microsoft}
Lin, T.Y., Maire, M., Belongie, S., Hays, J., Perona, P., Ramanan, D.,
  Doll{\'a}r, P., Zitnick, C.L.: Microsoft coco: Common objects in context. In:
  European conference on computer vision. pp. 740--755. Springer (2014)

\bibitem{liu2019image}
Liu, J., Sun, Y., Xu, X., Kamilov, U.S.: Image restoration using total
  variation regularized deep image prior. In: ICASSP 2019-2019 IEEE
  International Conference on Acoustics, Speech and Signal Processing (ICASSP).
  pp. 7715--7719. Ieee (2019)

\bibitem{NEURIPS2019_9015}
Paszke, A., Gross, S., Massa, F., Lerer, A., Bradbury, J., Chanan, G., Killeen,
  T., Lin, Z., Gimelshein, N., Antiga, L., Desmaison, A., Kopf, A., Yang, E.,
  DeVito, Z., Raison, M., Tejani, A., Chilamkurthy, S., Steiner, B., Fang, L.,
  Bai, J., Chintala, S.: Pytorch: An imperative style, high-performance deep
  learning library. In: Wallach, H., Larochelle, H., Beygelzimer, A.,
  d\textquotesingle Alch\'{e}-Buc, F., Fox, E., Garnett, R. (eds.) Advances in
  Neural Information Processing Systems 32, pp. 8024--8035. Curran Associates,
  Inc. (2019),
  \url{http://papers.neurips.cc/paper/9015-pytorch-an-imperative-style-high-performance-deep-learning-library.pdf}

\bibitem{punnappurath2019learning}
Punnappurath, A., Brown, M.S.: Learning raw image reconstruction-aware deep
  image compressors. IEEE transactions on pattern analysis and machine
  intelligence  \textbf{42}(4),  1013--1019 (2019)

\bibitem{schwartz2018deepisp}
Schwartz, E., Giryes, R., Bronstein, A.M.: Deepisp: Toward learning an
  end-to-end image processing pipeline. IEEE Transactions on Image Processing
  \textbf{28}(2),  912--923 (2018)

\bibitem{Shi_2016_CVPR}
Shi, W., Caballero, J., Huszar, F., Totz, J., Aitken, A.P., Bishop, R.,
  Rueckert, D., Wang, Z.: Real-time single image and video super-resolution
  using an efficient sub-pixel convolutional neural network. In: Proceedings of
  the IEEE Conference on Computer Vision and Pattern Recognition (CVPR) (June
  2016)

\bibitem{Simonyan15}
Simonyan, K., Zisserman, A.: Very deep convolutional networks for large-scale
  image recognition. In: International Conference on Learning Representations
  (2015)

\bibitem{wang2020multi}
Wang, J., Deng, X., Xu, M., Chen, C., Song, Y.: Multi-level wavelet-based
  generative adversarial network for perceptual quality enhancement of
  compressed video. In: European Conference on Computer Vision. pp. 405--421.
  Springer (2020)

\bibitem{1292216}
Wang, Z., Simoncelli, E., Bovik, A.: Multiscale structural similarity for image
  quality assessment. In: The Thrity-Seventh Asilomar Conference on Signals,
  Systems \& Computers, 2003. vol.~2, pp. 1398--1402 Vol.2 (2003).
  \doi{10.1109/ACSSC.2003.1292216}

\bibitem{xing2021invertible}
Xing, Y., Qian, Z., Chen, Q.: Invertible image signal processing. In:
  Proceedings of the IEEE/CVF Conference on Computer Vision and Pattern
  Recognition. pp. 6287--6296 (2021)

\bibitem{zamir2020cycleisp}
Zamir, S.W., Arora, A., Khan, S., Hayat, M., Khan, F.S., Yang, M.H., Shao, L.:
  Cycleisp: Real image restoration via improved data synthesis. In: Proceedings
  of the IEEE/CVF Conference on Computer Vision and Pattern Recognition. pp.
  2696--2705 (2020)

\end{thebibliography}
\end{document}